\documentclass[sigconf]{acmart}

\AtBeginDocument{%
  }

\setcopyright{acmlicensed}
\copyrightyear{2018}
\acmYear{2018}
\acmDOI{XXXXXXX.XXXXXXX}
\acmConference[Conference acronym 'XX]{Make sure to enter the correct
  conference title from your rights confirmation email}{June 03--05,
  2018}{Woodstock, NY}
\acmISBN{978-1-4503-XXXX-X/2018/06}

\usepackage{graphicx}
\usepackage{subcaption}
\usepackage{multirow}
\usepackage{fvextra}
\usepackage[table]{xcolor}
\usepackage{xcolor} 
\usepackage{booktabs}
\definecolor{linkblue}{RGB}{0, 70, 140}
\hypersetup{
    colorlinks=true,
    linkcolor=linkblue,   
    citecolor=linkblue,   
    urlcolor=linkblue     
}
\definecolor{lightblue}{RGB}{100,149,237}
\hypersetup{
    colorlinks=true,
    linkcolor=linkblue,   
    citecolor=linkblue,   
    urlcolor=linkblue     
}

\usepackage{tikz}

\newcommand{\circled}[1]{%
\tikz[baseline=(char.base)]{
\node[shape=circle, fill=black, text=white, 
inner sep=0.8pt, font=\scriptsize] (char) {#1};}}
\begin{document}

\title{Mitigating Entangled Steering in Large Vision-Language Models for Hallucination Reduction}

\author{Yuanhong Zhang}
\authornote{Both authors contributed equally to this research.}
\email{yuanhongzhang@stu.xjtu.edu.cn}
\affiliation{%
\institution{Ministry of Education Key Laboratory of Intelligent Networks and Network Security}
  \institution{Xi`an Jiaotong University}
  \city{Xi`an}
  \country{China}
}

\author{Zhaoyang Wang}
\authornotemark[1]
\affiliation{%
  \institution{Ministry of Education Key Laboratory of Intelligent Networks and Network Security}
  \institution{Xi`an Jiaotong University}
  \city{Xi`an}
  \country{China}}
\email{3124351019@stu.xjtu.edu.cn}

\author{Xin Zhang}
\affiliation{%
   \institution{Centre for Frontier AI Research, Institute of High Performance Computing, Agency for Science, Technology and Research}
  \country{Singapore}
}
\email{xinzhang_xd@163.com}

\author{Weizhan Zhang}
\authornote{Corresponding author.}
\affiliation{%
 \institution{Ministry of Education Key Laboratory of Intelligent Networks and Network Security}
  \institution{Xi`an Jiaotong University}
  \city{Xi`an}
  \country{China}}
\email{zhangwzh@xjtu.edu.cn}



\author{Joey Tianyi Zhou}
\affiliation{%
   \institution{Centre for Frontier AI Research, Institute of High Performance Computing, Agency for Science, Technology and Research}
  \country{Singapore}
}
\email{joey_zhou@a-star.edu.sg}




\begin{abstract}
Large Vision-Language Models (LVLMs) have achieved remarkable success across cross-modal tasks but remain hindered by hallucinations, producing textual outputs inconsistent with visual content. Existing methods mitigate hallucinations but often alter generation behavior, resulting in shorter outputs and shifted token distributions, especially in latent space steering approaches.
We identify that this issue stems from entangled steering signals, where suppressing hallucinations inadvertently disrupts the model’s intrinsic generation behavior.
To address this, we propose MESA, an effective plug-and-play framework that performs controlled and selective latent intervention for hallucination mitigation. 
Specifically, MESA targets hallucination-relevant responses while preserving the model’s original token distribution, enabling effective hallucination reduction without compromising generation behavior.
Extensive experiments across diverse generative and discriminative benchmarks demonstrate that MESA consistently reduces hallucinations while better preserving generation behavior, outperforming prior methods across multiple LVLM families.
\end{abstract}




\maketitle

\section{Introduction}
Recently, Large Vision-Language Models (LVLMs)~\cite{li2022blip,liu2024improved,liu2023visual,zhu2023minigpt,bai2023qwen,chen2024internvl,ye2024mplug} have demonstrated impressive performance in various tasks such as visual question answering~\cite{huynh2025visual,li2024configure,hartsock2024vision} and image captioning~\cite{dong2024benchmarking,li2023blip,hossain2019comprehensive}. By projecting visual signals into the language space, LVLMs are able to capture rich semantic information and express it in natural language~\cite{touvron2023llama,liu2023visual,zhang2024vision}. Despite these advances, they still
suffer from several challenges that limits their reliability and applicability in various real-world tasks~\cite{an2025agla,huang2025shield,chen2024detecting}. In particular, one prominent challenge lies with hallucinations, where LVLMs generate inaccurate or ungrounded descriptions, such as introducing non-existent objects or  misrepresenting relationships~\cite{bai2024hallucination,li2024surveying,liu2024survey}.

Existing approaches to mitigate hallucinations in LVLMs can be broadly divided into training-based and training-free methods. Training-based methods rely on high-quality data~\cite{liu2023mitigating,yu2024rlhf,yu2025rlaif} or preference optimization~\cite{ouali2024clip} to strengthen visual grounding, but incur significant human and computational costs~\cite{huang2025shield}. In contrast, training-free methods intervene at inference time via decoding penalties~\cite{leng2024vcd,wang2024icd}, latent space steering~\cite{yang2025nullu,liu2025vti}, or attention-based visual enhancement~\cite{an2025agla,yin2025clearsight,chen2025ict}.
In particular, recent studies~\cite{su2025activation,bai2024hallucination} show that hallucinations manifest in the hidden states of LVLMs during generation. This motivates modeling and intervening in the representation space as a more fundamental and interpretable approach to understanding and mitigating hallucinations~\cite{su2025activation,liu2025vti}.

Although the representation-level perspective has inspired latent space steering methods for hallucination mitigation~\cite{liu2025vti,yang2025nullu}, their effects remain insufficiently understood. For the first time, we identify a critical limitation: despite improving hallucination metrics, these methods do so at the cost of altering the generation process. Specifically, they lead to shorter outputs and systematically shifted token distributions in generative scenarios (see ~\autoref{fig:motivation_all}). This indicates that existing interventions fail to achieve selective hallucination control, instead entangling hallucination suppression with broader changes in generation behavior.

\begin{figure}[t!]
\centering
\includegraphics[width=0.49\textwidth]{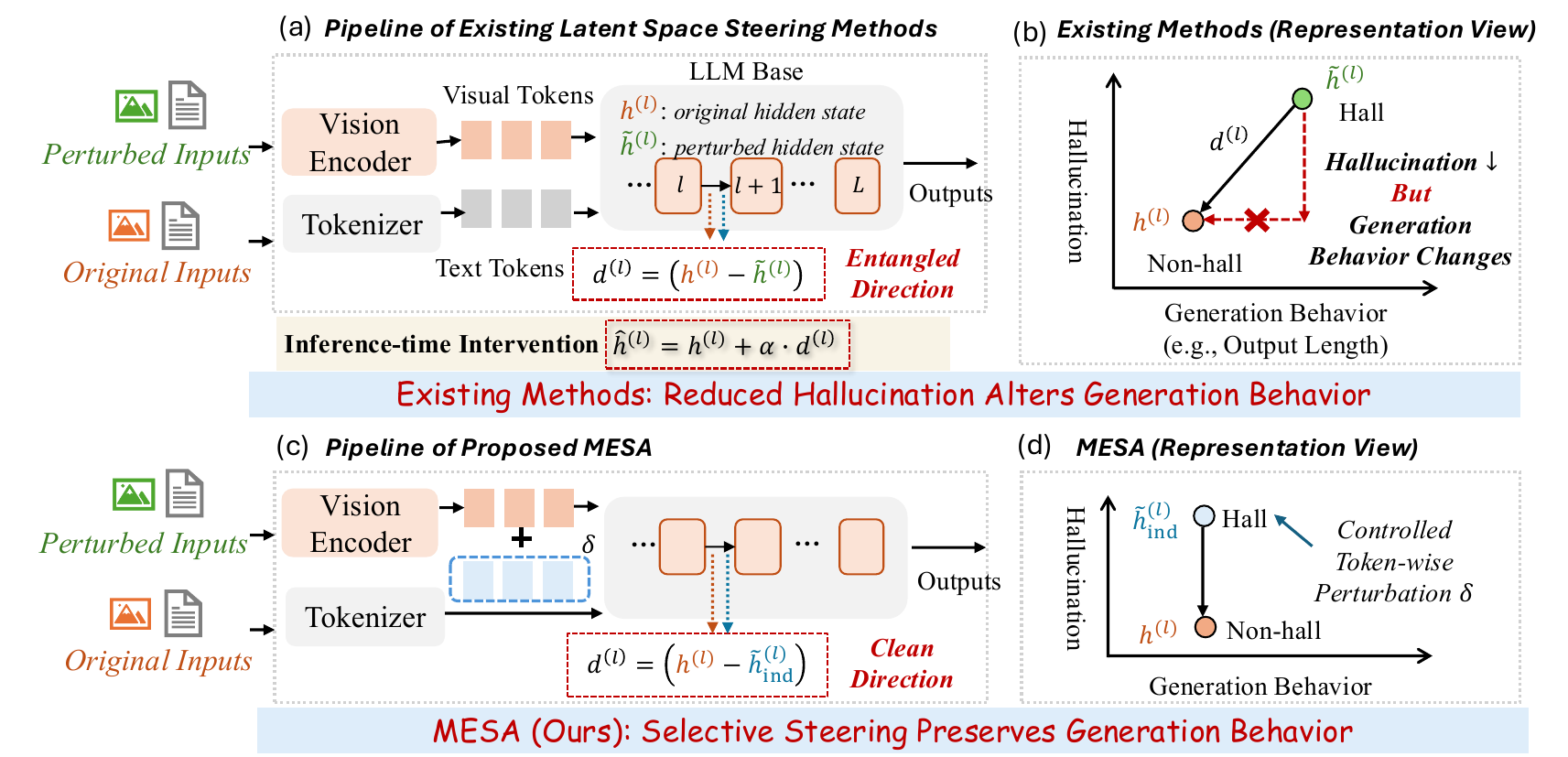}
\caption{Motivation of MESA. Existing methods produce entangled steering directions from stochastic perturbations, reducing hallucination but altering generation behavior (top).
MESA instead induces hallucination-specific representations via controlled perturbations, resulting in a clean steering direction that enables selective mitigation while preserving generation behavior (bottom). }
\label{fig:motivation_mesa}
\end{figure}

To address this issue, we propose MESA, a plug-and-play framework that mitigates entangled steering via controlled perturbations, reducing hallucinations while preserving generation behavior. 
We first analyze why existing latent space steering methods affect generation behavior. We find that the steering direction is inherently entangled, as it is constructed from contrastive representations between the original input and its perturbed counterpart. The stochastic perturbation introduces uncontrolled uncertainty, which simultaneously affects hallucination-related factors and the overall generation dynamics, leading to mixed and non-selective directions.
Based on this insight, MESA shifts from estimating steering directions to inducing hallucination-specific representations via controlled, token-wise perturbations on visual features, promoting hallucination-related behavior while preserving the overall generation process.
To learn such perturbations, we optimize two complementary objectives: one enhances hallucination-related responses, while the other preserves the original distribution. This yields cleaner hallucination representations and more selective steering, reducing hallucinations without compromising the model’s original capabilities.

Extensive experiments over multiple generative and discriminative benchmarks show that MESA achieves superior performance on hallucination mitigation as well as preserving original generation behaviors of LVLMs. Our contributions are summarized as:
\begin{itemize}
    \item We first reveal that latent space steering not only reduces hallucinations but also alters generation behavior, and identify entangled steering vectors as the underlying cause.
    \item We propose MESA, a plug-and-play framework that mitigates entangled steering by inducing cleaner hallucination-related directions through controlled perturbations with distribution-preserving constraints, thereby reducing hallucinations in LVLMs.
    \item Comprehensive experiments validate MESA’s effectiveness in mitigating hallucinations across diverse benchmarks and LVLM families. Moreover, it better preserves generation behavior compared to existing methods.
\end{itemize}

\section{Related Work}
\label{sec:related_work}

\subsection{Large Vision-Language Models (LVLMs)}
Benefiting from the success of large language models (LLMs)~\cite{gilardi2023chatgpt,touvron2023llama,chowdhery2023palm,chiang2023vicuna,bai2023qwen}, Large Vision-Language Models (LVLMs) have developed rapidly~\cite{li2022blip,li2023blip,liu2024improved,liu2023visual,zhu2023minigpt,bai2023qwen,chen2024internvl,ye2024mplug} in recent years.
Early works (e.g., BLIP~\cite{li2022blip}) typically bridge visual and textual modalities by linking a vision encoder to an LLM through dedicated alignment modules, such as linear projection layers or Q-formers~\cite{li2023blip}. Building on this foundation, subsequent works (e.g., LLaVA~\cite{liu2023visual}, Qwen-VL~\cite{bai2023qwen}) introduce visual instruction tuning, which further enhances LVLMs, enabling them to perform more advanced reasoning and achieve deeper multimodal understanding.
However, similar to LLMs, LVLMs inevitably suffer from hallucinations~\cite{bai2024hallucination}, which undermine their robustness and reliability in real-world applications.  

\subsection{Mitigating Hallucinations in LVLMs}
Approaches to mitigating hallucinations in LVLMs can be roughly divided into two categories based on the stage they take place~\cite{bai2024hallucination,chen2025ict}, i.e., training-based and training-free methods. 
At training stage, several research try to construct better data to train models~\cite{liu2023mitigating,yu2024rlhf,yu2025rlaif}. For example, LRA-instruction~\cite{liu2023mitigating} incorporates both positive and semantically structured negative dataset to reduce hallucinations, while RLHF-V~\cite{yu2024rlhf} proposes preference optimization to improve trustworthiness. However, despite their effectiveness, these methods are costly in terms of both time and computational resources.

In response, training-free strategies have attracted increasing attention~\cite{leng2024vcd,wang2024icd,yin2025clearsight,chen2025ict,yin2026dynamic,an2025agla, liu2025vti,yang2025nullu}. For instance, VCD~\cite{leng2024vcd} employs a contrastive decoding scheme that leverages both the original output and noise-perturbed output to mitigate hallucinations. However, such methods often compromise the quality of the generated content~\cite{chen2025ict}. 
Another line of work (e.g., VAF~\cite{yin2025clearsight}, ICT~\cite{chen2025ict}, and DMAS~\cite{yin2026dynamic}) addresses hallucinations by manipulating the attention weights assigned to visual inputs. 
In addition, some works (e.g., VTI~\cite{liu2025vti} and Nullu~\cite{yang2025nullu}) adjust model behavior by modifying features or weights based on steering directions derived from positive and negative samples, which can be broadly viewed as a form of latent space steering. 

\begin{figure*}[t!]
\centering

\begin{subfigure}[b]{0.24\textwidth}
    \centering
    \includegraphics[width=\textwidth]{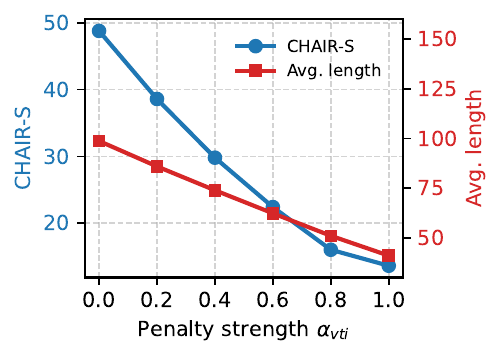}
    \caption{VTI}
    \label{fig:motivation1_vti}
\end{subfigure}
\hfill
\begin{subfigure}[b]{0.24\textwidth}
    \centering
    \includegraphics[width=\textwidth]{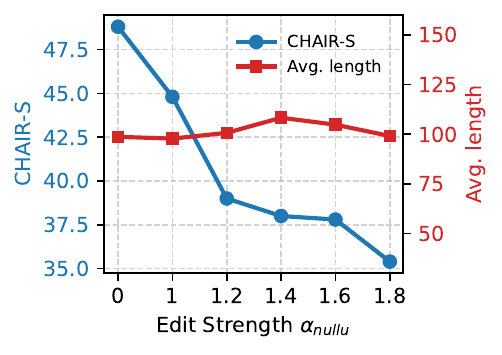}
    \caption{Nullu}
    \label{fig:motivation1_nullu}
\end{subfigure}
\hfill
\begin{subfigure}[b]{0.24\textwidth}
    \centering
    \includegraphics[width=\textwidth]{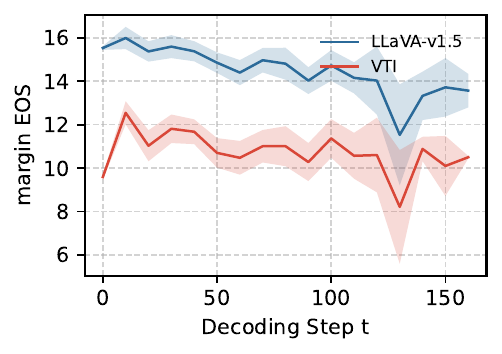}
    \caption{EOS margin (VTI)}
    \label{fig:motivation2_eos_vti}
\end{subfigure}
\hfill
\begin{subfigure}[b]{0.24\textwidth}
    \centering
    \includegraphics[width=\textwidth]{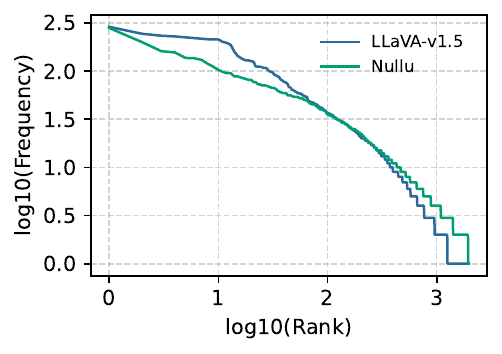}
    \caption{Zipf distribution (Nullu)}
    \label{fig:motivation3_zipf_nullu}
\end{subfigure}

\caption{
Analysis of latent space steering effects on hallucination and generation behavior.
(Left) Relationship between steering strength $\alpha$, hallucination ($CHAIR_S$), and output length for VTI and Nullu.
(Right) Mechanistic analysis: VTI shifts EOS competition, while Nullu alters token frequency distribution.
}
\label{fig:motivation_all}
\end{figure*}


\section{Preliminaries and Empirical Analysis}

\subsection{Latent Space Steering in LVLMs}
A large vision-language model (LVLM) parameterized by $\theta$ takes a visual sequence $V = \{v_1, \dots, v_{N_v}\}$ and a textual prompt $X = \{x_1, \dots, x_{N_x}\}$ as input, where $N_v$ and $N_x$ denote their respective lengths. Let $E_v(\cdot)$ and $E_x(\cdot)$ denote the visual and textual embedding functions, respectively. The inputs are embedded and concatenated into a unified sequence $h^{(0)} = \text{concat}(E_v(V), E_x(X))$, which is then processed through $L$ transformer layers:
\begin{equation}
h^{(l+1)} = \mathcal{F}^{(l)}(h^{(l)}), \quad l = 0, \dots, L-1,
\label{eq:transformer}
\end{equation}
where $\mathcal{F}^{(l)}(\cdot)$ denotes the $l$-th transformer block. The hidden states $h^{(l)} \in \mathbb{R}^{N \times D}$ represent the joint visual-text token sequence with $N = N_v + N_x$, and $h_i^{(l)} \in \mathbb{R}^{D}$ denotes the $i$-th token. At each decoding step $t$, the next token is predicted from the final-layer hidden state $h^{(L)}$, producing logits $z_t$.

Building upon the hidden representations $h^{(l)}$ defined above, recent approaches~\cite{liu2025vti,yang2025nullu} mitigate hallucination via latent space steering, which intervenes in these representations during inference process. Specifically, at layer $l$, a steering direction $d^{(l)} \in \mathbb{R}^{D}$ is added as:
\begin{equation}
\hat{h}^{(l)} = h^{(l)} + \alpha d^{(l)},
\label{eq:steer}
\end{equation}
where $d^{(l)}$ is broadcast across all tokens and $\alpha$ controls intervention strength. The direction $d^{(l)}$ encodes a specific behavioral shift and is typically constructed by contrasting representations under different conditions (e.g., original versus perturbed inputs~\cite{leng2024vcd,liu2025vti}).

\subsection{Empirical Analysis of Steering Direction Entanglement}
\label{sec:motivation}
In this section, we systematically investigate whether latent space steering can selectively control hallucination. Contrary to the common assumption that steering enables independent manipulation of hallucination~\cite{liu2025vti,su2025activation,bai2024hallucination}, we reveal a fundamental limitation: \textit{steering directions are inherently entangled with generation behavior}. 

\circled{1} \ \textbf{Hallucination Reduction via Steering Affects Generation Behavior.}
As shown in ~\autoref{fig:motivation_all}, we vary the steering strength $\alpha$ to jointly evaluate hallucination (e.g., $CHAIR_S$) and generation behavior (e.g., output length and token distribution) using latent space steering methods such as VTI~\cite{liu2025vti} and Nullu~\cite{yang2025nullu}. We observe that increasing $\alpha$ in VTI reduces hallucination but progressively shortens the outputs (~\autoref{fig:motivation1_vti}). In contrast, Nullu maintains or slightly increases output length (~\autoref{fig:motivation1_nullu}), while inducing noticeable shifts in token distribution, resulting in more complex but less controlled generation.
These findings demonstrate that latent space steering does not achieve selective hallucination control. Instead, it alters the underlying generation process, where hallucination reduction emerges as a side effect of broader behavioral changes. 
In the following, we further analyze this entanglement from two perspectives: generation termination and token distribution.


\begin{figure*}[t!]
\centering
\includegraphics[width=0.85\textwidth]{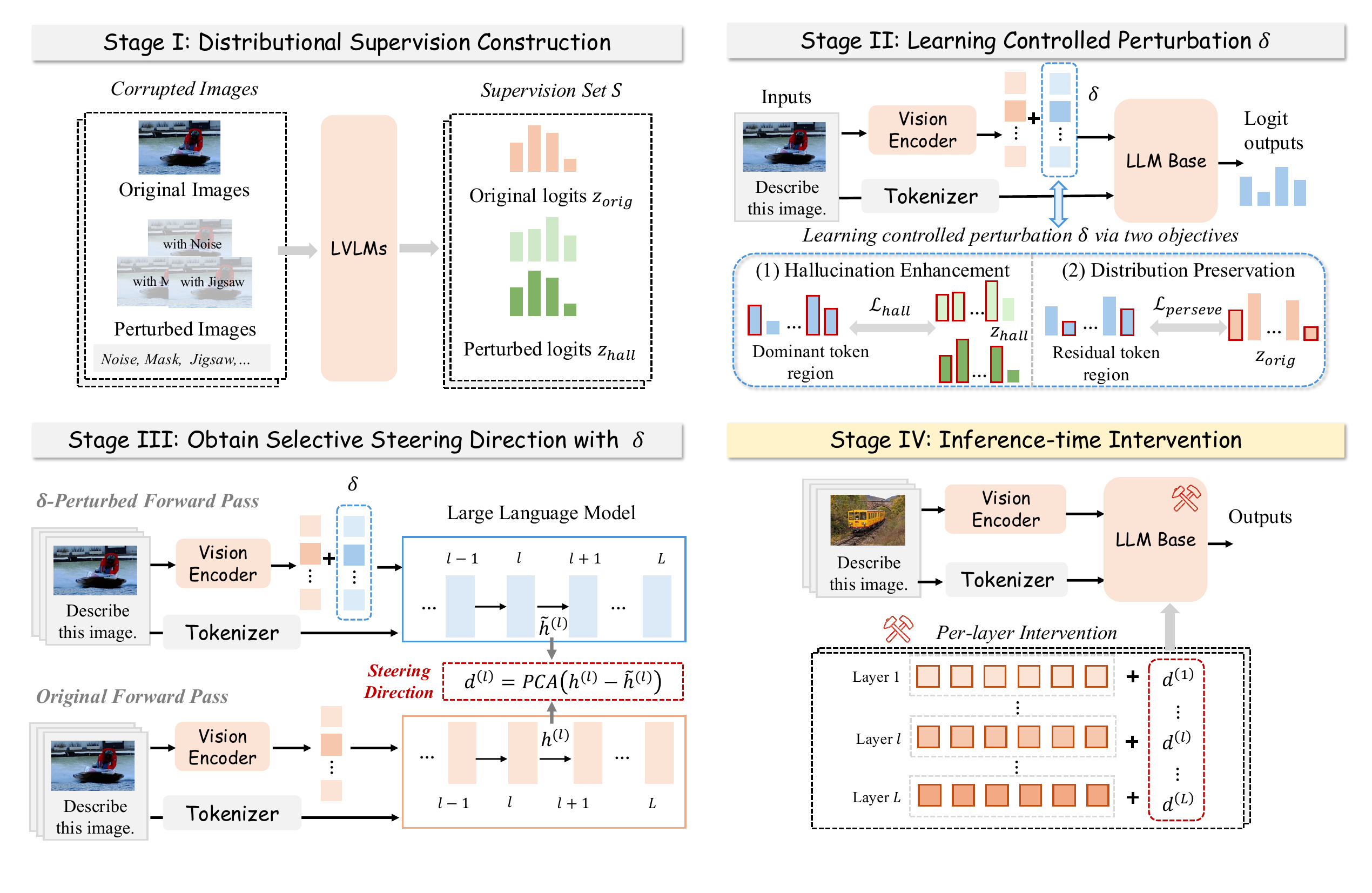}
\caption{Overview of MESA. MESA decomposes hallucination mitigation into three offline stages and an inference-time intervention. We first construct a distributional supervision set from original and degraded inputs (Stage I), then learn a controlled perturbation to induce hallucination-aware representations while preserving semantics (Stage II), and extract steering directions via PCA over induced representation shifts (Stage III). At inference, the learned directions are injected into hidden states to selectively correct hallucination without disrupting generation dynamics (Stage IV). }
\label{fig:overview_mesa}
\end{figure*}

\circled{2} \ \textbf{Steering Promotes Early Termination via EOS Bias.}
To understand why such steering methods (i.e., VTI) leads to shorter outputs, we analyze its effect on the generation termination mechanism.
We fix the generation trajectory via teacher forcing and examine the competition between the EOS token and non-EOS tokens using the EOS margin, defined as:
\begin{equation}
\text{margin}_{\text{EOS}}^{t} = \max_{y \neq \text{EOS}} z_t(y) - z_t(\text{EOS}),
\label{eq:eos}
\end{equation}
where $z_t(\cdot)$ denotes the logit at decoding step $t$.
As shown in ~\autoref{fig:motivation2_eos_vti}, the EOS margin under VTI is consistently lower than the baseline across decoding steps, making EOS more likely to be selected and leading to earlier termination and shorter outputs. This suggests that VTI reduces hallucination by promoting early termination rather than selectively correcting hallucinated content.

\circled{3} \ \textbf{Steering Induces Token Distribution Shift.}
Unlike VTI, Nullu~\cite{yang2025nullu} does not reduce output length and may even produce slightly longer responses (see ~\autoref{fig:motivation1_nullu}), suggesting a shift toward more complex generation.
To further investigate this behavior, we analyze the token frequency distribution using Zipf’s law~\cite{mikhaylovskiy2025zipf}, which characterizes the relationship between token rank and frequency. As shown in ~\autoref{fig:motivation3_zipf_nullu}, the x-axis represents the log-scaled token rank, while the y-axis denotes the log-scaled frequency. Each curve reflects the overall distribution of generated tokens.
Compared to the baseline, Nullu yields a flatter Zipf curve with a pronounced tail elevation, indicating increased low-frequency token usage and reduced dominance of high-frequency ones~\cite{mikhaylovskiy2025zipf}. While this may suggest higher lexical diversity, it also reflects a deviation from the base model’s intrinsic distribution, potentially introducing higher variance, over-specific predictions, and increased hallucination risk~\cite{holtzman2019curious,ji2023survey}. These observations suggest that, despite subspace-based decomposition to remove hallucination components, Nullu still alters generation behavior, particularly in token distribution.

In contrast, our proposed MESA further analyzes the underlying causes of this entanglement and mitigates hallucination while preserving generation behavior.

\section{MESA: Mitigating Entangled Steering for Hallucination Reduction}


\subsection{Problem Reformulation: Selective Steering}
Existing latent steering methods construct steering directions by introducing stochastic perturbations to the visual input~\cite{liu2025vti,chen2025ict}. Specifically, a perturbed input $\tilde{V}$ is obtained by applying random perturbations (e.g., Gaussian noise~\cite{leng2024vcd,chen2025ict} or masking~\cite{liu2025vti}) to the original image $V$. Given the same $X$, the steering direction is $d^{(l)} = h^{(l)}(V, X) - h^{(l)}(\tilde{V}, X)$. To analyze its limitation, we further decompose this change as:
\begin{equation}
d^{(l)} = h^{(l)}(V, X) - h^{(l)}(\tilde{V}, X)
= \Delta_{\text{hall}}^{(l)}
+ \Delta_{\text{non-hall}}^{(l)},
\end{equation}
where $\Delta_{\text{hall}}^{(l)}$ captures hallucination-related variation, while $\Delta_{\text{non-hall}}^{(l)}$ reflects changes in the overall generation process (as shown in ~\autoref{fig:motivation_mesa}), including token distribution and termination dynamics. Ideally, during inference, hallucination-related components should be adjusted from incorrect to grounded predictions, while non-hallucination factors such as generation behavior remain unchanged.

However, under existing stochastic perturbations~\cite{leng2024vcd,chen2025ict,an2025agla}, $\tilde{V}$ introduces uncontrolled uncertainty that simultaneously affects both components, preventing the resulting signal from being a clean, hallucination-specific representation. As a result, the estimated direction $d^{(l)}$ is inherently entangled, making selective hallucination control impossible without altering generation behavior, as observed in Sec.~\ref{sec:motivation}. These findings suggest that latent steering should be framed as a selective control problem, aiming to isolate $\Delta_{\text{hall}}^{(l)}$ while minimizing $\Delta_{\text{non-hall}}^{(l)}$. The optimization objective is:
\begin{equation}
\delta^* = \arg\max_{\delta} \ \Delta_{\text{hall}}^{(l)}-\; \lambda \, \Delta_{\text{non-hall}}^{(l)}
\end{equation}
where $\lambda$ is a trade-off hyperparameter, set to 1 in this paper. As illustrated in ~\autoref{fig:motivation_mesa}, instead of perturbing the input, we learn a perturbation feature $\delta$ applied to the visual tokens, which induces a hallucination-prone representation. 
This reformulation shifts latent steering from passively observing entangled differences to actively inducing hallucination-specific representation changes, selectively amplifying hallucination-related components while minimizing changes to non-hallucination factors.

\subsection{Constructing Steering Directions via Controlled Perturbations}

In this section, we describe how to obtain an offline steering direction by learning a controlled perturbation that induces hallucination-related behaviors and can be leveraged for mitigation. The framework consists of three components. First, we construct distributional supervision by collecting model outputs under original and degraded visual inputs, forming behaviorally meaningful target distributions. Second, we learn a conditional perturbation module that reshapes the model’s output distribution via alignment with these targets. Third, based on the learned perturbation, we derive a selective steering direction. The pipeline is illustrated in ~\autoref{fig:overview_mesa}.

\subsubsection{Stage I: Distributional Supervision Construction.}
Our goal is to learn a controlled, input-conditioned perturbation $\delta$ that selectively induces hallucination-related behaviors while preserving the original semantic distribution of the LVLM. To achieve this, we introduce a learnable, token-wise perturbation module $f_{\phi}$ (parameterized by $\phi$), which operates on the visual embeddings produced by the embedding function $E_v(\cdot)$. Formally, we define:
\begin{equation}
\delta = \operatorname{clip}\big(f_{\phi}(E_v(V)), -\epsilon, \epsilon\big),
\label{eq:delta}
\end{equation}
where $\operatorname{clip}(\cdot)$ denotes element-wise clipping, and $\epsilon$ controls the magnitude of the perturbation, ensuring it remains local and well-bounded. Instead of modifying the embedding function itself, we add $\delta$ to the visual tokens $E_v(V)$ and feed the perturbed sequence into the frozen transformer, yielding induced hidden states $\tilde{h}^{(l)}$ and logits $\tilde{z}_t$.

To supervise the optimization of $\phi$ while keeping the base model $\theta$ frozen, we construct two distinct logit-level signals: First, for the hallucination-inducing signals, we apply a set of visual degradations $\mathcal{T}=\{\tau_1,\dots,\tau_K\}$ (e.g., masking or blurring) to the input, yielding $V^{(k)}=\tau_k(V)$. The corresponding logits, denoted as $z_{\text{hall}}^{(k)}$, capture prior-driven behaviors under weakened visual input and serve as targets for hallucination induction. Second, for semantics-preserving signals, we use the baseline logits $z_{\text{orig}}$ from the original input, which capture the grounded predictive distribution and act as an anchor to prevent $\delta$ from distorting semantic coherence. 
Overall, the supervision set can be defined as:
\begin{equation}
\mathcal{S} = \{ z_{\text{orig}}, \{z_{\text{hall}}^{(k)}\}_{k=1}^K \}
\label{eq:supversion_set}
\end{equation}
thus provides a structured distributional target.

\subsubsection{Stage II: Learning Controlled Perturbations via Distribution Alignment.}
Given the constructed supervision set $\mathcal{S}$, we optimize the perturbation module $f_{\phi}$ by aligning the induced distribution with hallucination-inducing signals while preserving the original semantic structure.
Let $P_{\text{ind}} = \mathrm{softmax}(\tilde{z}_t)$ denote the induced token distribution produced by the perturbation module $f_{\phi}$. 
Additionally, let $P_{\text{hall}}^{(k)} = \mathrm{softmax}(z_{\text{hall}}^{(k)})$ and $P_{\text{orig}} = \mathrm{softmax}(z_{\text{orig}})$ denote the hallucination and original distributions, respectively. 

First, to induce hallucination behaviors, we align the induced distribution with the hallucination supervision set $\{z_{\text{hall}}^{(k)}\}_{k=1}^{K}$:
\begin{equation}
\mathcal{L}_{\text{hall}} =
\textstyle \sum_{k=1}^K w_k \cdot
D_{\mathrm{KL}}\big(
P_{\text{ind}} \,\|\, P_{\text{hall}}^{(k)}
\big)
\end{equation}
In practice, the divergence in $\mathcal{L}_{\text{hall}}$ is computed over a top-$m$ token subset to reduce low-probability noise and improve stability, as full-vocabulary optimization often results in unstable training and poor convergence~\cite{hoscilowicz2025adversarial}. In addition, the $w_k$ is a normalized weight. In particular, we adopt a dynamic weighting scheme: 
\begin{equation}
w_k \propto D_{\mathrm{KL}}\big(P_{\text{hall}}^{(k)} \,\|\, P_{\text{orig}}\big), 
\quad \textstyle \sum_{k=1}^{K} w_k = 1,
\label{eq:dynamic_weight}
\end{equation}
This design assigns larger weights to perturbations that induce stronger deviations from the original distribution, thereby prioritizing more challenging hallucination-inducing cases.

Second, to preserve semantic consistency, we constrain the induced distribution using the original supervision signal:
\begin{equation}
\mathcal{L}_{\text{preserve}} =
D_{\mathrm{KL}}\big(
P_{\text{ind}}
\,\|\, 
P_{\text{orig}}
\big).
\label{eq:dynamic_weight}
\end{equation}
In contrast to $\mathcal{L}_{\text{hall}}$, the divergence in $\mathcal{L}_{\text{preserve}}$ is computed over the top-$m$ tokens excluding the highest-probability ones (e.g., top-5). 
This design allows high-probability tokens to be freely adjusted, as hallucinations typically arise in them, while constraining the remaining tokens to preserve the overall distribution, equivalent to minimizing $\Delta_{\text{non-hall}}^{(l)}$ and maintaining semantic consistency and fluent generation.

The overall training objective for perturbation $\delta$ is as follows:
\begin{equation}
\mathcal{L} = \mathcal{L}_{\text{hall}} + \mathcal{L}_{\text{preserve}},
\label{eq:loss_total}
\end{equation}
The hallucination alignment term $\mathcal{L}_{\text{hall}}$ encourages the induced distribution to capture prior-driven behaviors under degraded inputs, while the preservation term $\mathcal{L}_{\text{preserve}}$ prevents semantic distribution drift. Together, these objectives optimize $f_{\phi}$ to produce controlled perturbations $\delta$ that shape the induced hidden states $\tilde{h}^{(l)}$ into hallucination-specific directions for subsequent steering, while preserving semantic consistency.

\subsubsection{Stage III: Obtain Selective Steering Direction}
After learning the token-wise perturbation $\delta$, we apply it to the projected visual embeddings to obtain the hallucination-induced embeddings (see ~\autoref{eq:delta}). These embeddings are then concatenated with the textual embeddings and fed into the $L$-layer transformer, yielding induced hidden representations $\tilde{h}^{(l)}$ at layer $l$, with the original representations denoted as $h^{(l)}$.
We compute $\Delta h^{(l)} = h^{(l)}-\tilde{h}^{(l)}$ at each layer, capturing perturbation-induced shifts in the representation space. PCA~\cite{mackiewicz1993principal} is then applied to $\Delta h^{(l)}$ across samples to extract the dominant direction for steering direction $d^{(l)}$~\cite{yang2025nullu,liu2025vti}.

\subsection{Inference-Time Intervention}

In the Stage IV, at inference time, we apply the learned steering direction to intervene in the generation process without modifying model parameters. Specifically, for each decoding step, we shift the hidden representations of the text decoder at layer $l$ using the direction $d^{(l)}$ as defined in ~\autoref{eq:steer}. 
The intervention is applied to hidden states of generated tokens, enabling controlled behavior modification during generation. Its strength is governed by a scaling factor $\alpha$, balancing hallucination mitigation and preservation of the original behavior. By injecting the steering direction into the latent space, the method selectively suppresses hallucination while maintaining overall coherence and distribution.

\begin{table*}[t!]
\centering
\small
\renewcommand{\arraystretch}{1.1}
\setlength{\tabcolsep}{7pt}
\caption{CHAIR hallucination evaluation results on different LVLMs with max new tokens set to 512.}
\begin{tabular}{c cccc cccc}
\toprule
\multirow{2}{*}{Method}
& \multicolumn{4}{c}{\textbf{LLaVA-v1.5}}
& \multicolumn{4}{c}{\textbf{Qwen-VL}} \\

\cmidrule(lr){2-5} \cmidrule(lr){6-9}

& $\mathrm{CHAIR}_S$\scriptsize$\downarrow$ 
& $\mathrm{CHAIR}_I$\scriptsize$\downarrow$ 
& Recall\scriptsize$\uparrow$ 
& Avg. length\scriptsize$\uparrow$
& $\mathrm{CHAIR}_S$\scriptsize$\downarrow$ 
& $\mathrm{CHAIR}_I$\scriptsize$\downarrow$ 
& Recall\scriptsize$\uparrow$ 
& Avg. length\scriptsize$\uparrow$ \\
\midrule

Vanilla & 48.80 & 13.40 & 77.70 & 98.70 & 38.70 & 14.30 & 72.70 & 82.40 \\
VCD~\cite{leng2024vcd}    & 56.30 & 16.00 & 77.10 & 99.60 & 37.60 & 13.20 & 64.90 & 71.30 \\
ICD~\cite{wang2024icd}     & 59.50 & 15.80 & 76.30 & \textbf{103.7} & 49.90 & 13.80 & 71.90 & \textbf{101.1} \\
VAF~\cite{yin2025clearsight}     & 51.10 & 14.00 & \textbf{77.90} & 96.60 & 36.90 & 10.60 & 72.50 & 79.50 \\
ICT~\cite{chen2025ict}     & 51.00 & 14.90 & 76.70 & 101.20 & 38.10 & 11.40 & \textbf{74.40} & 85.10 \\
VTI~\cite{liu2025vti}     & 40.70 & 12.00 & 75.10 & 80.30 & 31.40 & 9.30 & 72.30 & 70.10 \\
Nullu~\cite{yang2025nullu}   & 44.80 & 13.00 & 75.10 & 97.80 & 34.20 & 10.30 & 71.80 & 81.00 \\
\rowcolor{gray!15}
\textbf{MESA} & \textbf{31.00} & \textbf{8.60} & 75.80 & 93.50 & \textbf{29.00} & \textbf{8.20} & 72.70 & 83.70 \\
\bottomrule
\end{tabular}
\label{tab:chair}
\end{table*}

\begin{table*}[t!]
\centering
\caption{Comparison results on the POPE benchmark. Results are reported in terms of Accuracy (\%) and F1 Score (\%). The proposed method achieves consistent improvements over related methods, with the best results highlighted in bold.}
\small
\setlength{\tabcolsep}{3.5pt}

\begin{tabular}{c|c|cccccccccccc}
\toprule
\multirow{3}{*}{Model} & \multirow{3}{*}{Method} 
& \multicolumn{6}{c}{\textbf{MSCOCO}} 
& \multicolumn{6}{c}{\textbf{GQA}} \\

\cmidrule(l){3-8} \cmidrule(l){9-14}
& 
& \multicolumn{2}{c}{Random} 
& \multicolumn{2}{c}{Popular} 
& \multicolumn{2}{c}{Adversarial}
& \multicolumn{2}{c}{Random} 
& \multicolumn{2}{c}{Popular} 
& \multicolumn{2}{c}{Adversarial} \\

\cmidrule(l){3-14}
& 
& Accuracy$\uparrow$ & F1$\uparrow$  
& Accuracy$\uparrow$ & F1$\uparrow$ 
& Accuracy$\uparrow$ & F1$\uparrow$  
& Accuracy$\uparrow$ & F1$\uparrow$ 
& Accuracy$\uparrow$ & F1$\uparrow$ 
& Accuracy$\uparrow$ & F1$\uparrow$   \\
\midrule

\multirow{8}{*}{\textbf{LLaVA-v1.5}}
& Vanilla & 85.53 & 88.80 & 83.20  & 84.28 & 75.60 & 78.63 & 84.96 & 86.25 & 73.30 & 77.93 & 67.4 & 74.31 \\
& VCD~\cite{leng2024vcd}    & 87.73 & 87.16 & 85.38 & 85.06 & 80.88 & 81.33 & 86.65 & 86.99 & 80.73 & 82.24 & 76.09 & 78.78 \\
& ICD~\cite{wang2024icd}    & 85.23 & 83.52 & 83.73 & 82.00 & 81.46 & 80.32 & 85.90 & 85.29 & 78.56 & 78.79 & 75.90 & 76.89 \\
& VAF~\cite{yin2025clearsight}    & 87.36 & 85.94 & 86.46 & 85.09 & \textbf{83.86} & 82.71 & 89.00 & 88.63 & 81.16 & 82.00 & 79.10 & 80.41 \\
& ICT~\cite{chen2025ict}    & 88.96 & 87.85 & 85.86 & 84.36 & 83.83 & 82.45 & 89.26 & 88.78 & 82.86 & 83.70 & 79.13 & 81.07 \\
& VTI~\cite{liu2025vti}    & 87.70 & 86.45 & 86.37 & 85.20 & 83.43 & 82.54 & 88.37 & 88.02 & 81.73 & 82.39 & 78.33 & 79.78 \\
& Nullu~\cite{yang2025nullu}  & 89.03 & 88.52 & 86.06 & 85.88 & 78.66 & 79.57 & 87.70 & 88.00 & 77.63 & 80.13 & 72.10 & 76.37 \\
& \cellcolor{gray!15}\textbf{MESA}   
& \cellcolor{gray!15}\textbf{90.27} 
& \cellcolor{gray!15}\textbf{90.20} 
& \cellcolor{gray!15}\textbf{87.63} 
& \cellcolor{gray!15}\textbf{87.32} 
& \cellcolor{gray!15}83.23 
& \cellcolor{gray!15}\textbf{83.56} 
& \cellcolor{gray!15}\textbf{89.50} 
& \cellcolor{gray!15}\textbf{89.45} 
& \cellcolor{gray!15}\textbf{82.93} 
& \cellcolor{gray!15}\textbf{83.91} 
& \cellcolor{gray!15}\textbf{79.60} 
& \cellcolor{gray!15}\textbf{81.35} \\
\midrule

\multirow{8}{*}{\textbf{Qwen-VL}}
& Vanilla & 87.23 & 85.50 & 85.10 & 84.78 & 78.63 & 81.44 & 74.76 & 66.60 & 74.76 & 66.66 & 73.53 & 65.53 \\
& VCD~\cite{leng2024vcd}    & 88.63 & \textbf{87.81} & 87.12 & 86.40 & 84.26 & 83.90 & 85.59 & 85.33 & 81.83 & 82.23 & 80.01 & 80.75 \\
& ICD~\cite{wang2024icd}    & 82.43 & 78.89 & 81.90 & 78.20 & 80.80 & 77.23 & 86.20 & 84.62 & 80.80 & 77.89 & 81.00 & 80.06 \\
& VAF~\cite{yin2025clearsight}    & 86.80 & 84.98 & 85.63 & 83.87 & 83.70 & 82.09 & 86.11 & 86.11 & 85.43 & 81.35 & 80.03 & 79.75 \\
& ICT~\cite{chen2025ict}    & 88.06 & 86.55 & 87.03 & 86.15 & 84.30 & 83.51 & 85.33 & 86.12 & 85.31 & 84.15 & 81.61 & 80.75 \\
& VTI~\cite{liu2025vti}    & 87.97 & 86.79 & 86.37 & 85.29 & 83.53 & 82.75 & 84.33 & 81.94 & 83.67 & 81.31 & 80.47 & 78.44 \\
& Nullu~\cite{yang2025nullu}  &88.26 & 86.13 & 85.43 & 85.21 & 84.11 & 82.09 & 74.73 & 65.55 & 74.71 & 66.54 & 73.61 & 65.59 \\
& \cellcolor{gray!15}\textbf{MESA}    
& \cellcolor{gray!15}\textbf{88.83} 
& \cellcolor{gray!15}87.11 
& \cellcolor{gray!15}\textbf{87.20} 
& \cellcolor{gray!15}\textbf{86.51} 
& \cellcolor{gray!15}\textbf{84.47} 
& \cellcolor{gray!15}\textbf{84.10}
& \cellcolor{gray!15}\textbf{87.97} 
& \cellcolor{gray!15}\textbf{86.67} 
& \cellcolor{gray!15}\textbf{86.07} 
& \cellcolor{gray!15}\textbf{84.89} 
& \cellcolor{gray!15}\textbf{82.73} 
& \cellcolor{gray!15}\textbf{81.93} \\
\bottomrule
\end{tabular}
\label{tab:pope}
\end{table*}

\begin{table}[t!]
\centering
\caption{Results on AMBER on LLaVA-v1.5.}
\setlength{\tabcolsep}{4pt}
\begin{tabular}{l|cccc}
\toprule
\multicolumn{5}{c}{\textbf{Generative Task}} \\
\cmidrule(lr){1-5}
Method & CHAIR\scriptsize$\downarrow$ & Cover\scriptsize$\uparrow$ & Hallucination\scriptsize$\downarrow$ & Cognition\scriptsize$\downarrow$ \\
\midrule
Vanilla       & 10.6 & 50.9 & 36.4 & 4.1 \\
VTI~\cite{liu2025vti}           & 6.7  & 47.6 & 28.1 & 3.6 \\
Nullu~\cite{yang2025nullu}         & 7.4  & 48.2 & 30.5 & 3.7 \\
\rowcolor{gray!15}
\textbf{MESA} & \textbf{6.4} & \textbf{50.4} & \textbf{27.4} & \textbf{3.2} \\
\midrule
\multicolumn{5}{c}{\textbf{Discriminative Task}} \\
\midrule
Method & Accuracy\scriptsize$\uparrow$ & Precision\scriptsize$\uparrow$ & Recall\scriptsize$\uparrow$ & F1\scriptsize$\uparrow$ \\
\midrule
Vanilla       & 71.4 & 81.9 & 73.1 & 77.2 \\
VTI~\cite{liu2025vti}           & 79.1 & 82.1 & \textbf{87.6} & 84.8 \\
Nullu~\cite{yang2025nullu}         & 81.5 & 87.4 & 84.2 & 85.8 \\
\rowcolor{gray!15}
\textbf{MESA} & \textbf{84.3} & \textbf{90.3} & 85.4 & \textbf{87.8} \\
\bottomrule
\end{tabular}
\label{tab:amber}
\end{table}

\section{Experiments}

\subsection{Experimental Settings.}

\subsubsection{Datasets and Metrics.}
We evaluate our model on both discriminative and generative datasets, as listed below. More details about the datasets are provided in the supplementary materials. 
(1) \textbf{CHAIR}: The dataset~\cite{rohrbach2018object} quantifies the object hallucinations in the image captioning tasks by comparing the generated objects to ground-truth objects. We randomly select 1000 images from the MSCOCO dataset~\cite{lin2014microsoft} and use $CHAIR_I$, $CHAIR_S$, Recall, and Average generated length as evaluation metrics. 
(2) \textbf{POPE}: The dataset~\cite{li2023evaluating} evaluates existence-level hallucinations in LVLMs. It contains 27,000 Yes/No questions based on object annotations in three datasets (i.e., MSCOCO~\cite{lin2014microsoft}, A-OKVQA~\cite{schwenk2022okvqa}, GQA~\cite{hudson2019gqa}), with metrics including Accuracy, Precision, Recall, and F1 score. It encompasses three sampling settings: random, popular, and adversarial. 
(3) \textbf{AMBER}: The dataset~\cite{wang2023amber} offers a comprehensive suite of generative and discriminative metrics, explicitly covering diverse hallucinations (e.g., existence, attribute, and relation).
(4) \textbf{LLaVA-Bench}: This generative dataset~\cite{liu2023visual} comprises 24 images and 60 questions, designed to evaluate LVLMs on challenging tasks and their generalization to new domains. 
Following prior work~\cite{an2025agla,yang2025nullu}, we use GPT-4o to assess the accuracy and informativeness of the generated captions.

\begin{figure}[t!]
\centering
\includegraphics[width=0.49\textwidth]{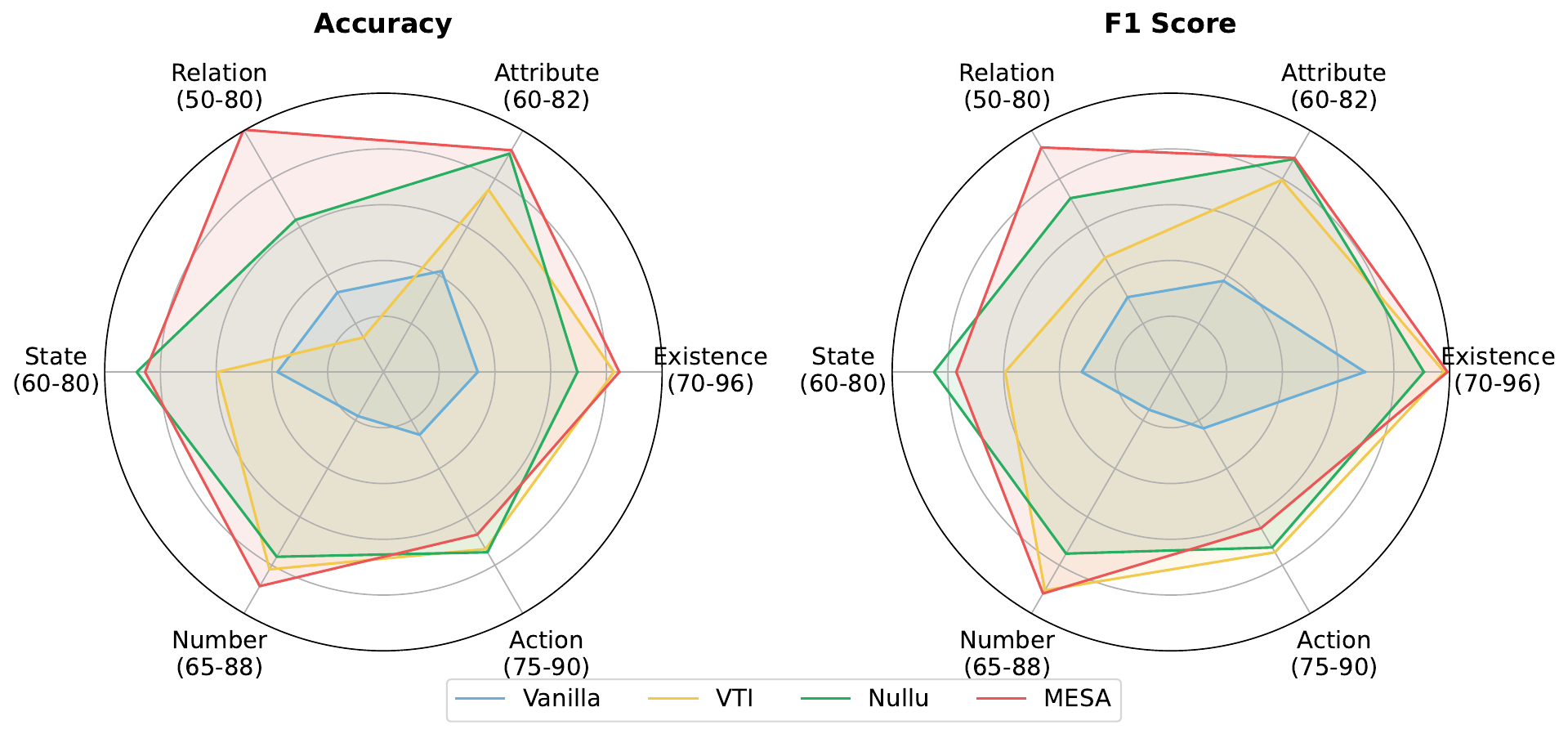}
\caption{Evaluation results of LLaVA-v1.5 on the AMBER benchmark for the discriminative task.}
\label{fig:radar_amber}
\end{figure}

\begin{table}[t]
\centering
\caption{Results on LLaVA-Bench. }
\setlength{\tabcolsep}{4pt}
\begin{tabular}{c|cc|cc|cc}
\toprule
Metric & Vanilla & MESA & VTI~\cite{liu2025vti} & MESA & Nullu~\cite{yang2025nullu} & MESA \\
\midrule
Acc $\uparrow$ 
& 6.26 & \textbf{6.43} 
& 3.93 & \textbf{5.67} 
& 6.10 & \textbf{6.52} \\

Det $\uparrow$ 
& 5.80 & \textbf{5.81} 
& 3.53 & \textbf{5.61} 
& 5.67 & \textbf{5.77} \\
\bottomrule
\end{tabular}
\label{tab:llava_bench}
\end{table}

\begin{table}[t!]
\centering
\caption{Ablation of training loss and perturbation designs.}
\setlength{\tabcolsep}{4pt}
\begin{tabular}{l|cccc}
\toprule
\multicolumn{5}{c}{\textbf{Training Loss for Perturbation}} \\
\cmidrule(lr){1-5}
Method        & $\mathrm{CHAIR}_S$\scriptsize$\downarrow$ 
& $\mathrm{CHAIR}_I$\scriptsize$\downarrow$ 
& Recall\scriptsize$\uparrow$ 
& Avg. length\scriptsize$\uparrow$ \\ 
\midrule
Vanilla       & 48.8 & 13.4 & 77.7          & 98.7  \\
w $\mathcal{L}_{\text{preserve}}$    & 41.6 & 11.2 & 75.6          & 93.8  \\
w $\mathcal{L}_{\text{hall}}$   & 46.4 & 12.2 & \textbf{79.3} & \textbf{104.4} \\
\cellcolor{gray!15}\textbf{MESA} & \cellcolor{gray!15}\textbf{31.0} & \cellcolor{gray!15}\textbf{8.6} & \cellcolor{gray!15}75.8 & \cellcolor{gray!15}93.5 \\
\midrule

\multicolumn{5}{c}{\textbf{Different Type of Perturbation}} \\
\cmidrule(lr){1-5}
Method        & $\mathrm{CHAIR}_S$\scriptsize$\downarrow$ 
& $\mathrm{CHAIR}_I$\scriptsize$\downarrow$ 
& Recall\scriptsize$\uparrow$ 
& Avg. length\scriptsize$\uparrow$ \\ 
\midrule
Vanilla       & 48.8 & 13.4 & \textbf{77.7} & 98.7  \\
Less perturb. & 41.4 & 11.3 & 76.6          & 99.6  \\
More perturb. & 39.2 & 10.1 & 75.8          & \textbf{106.4} \\
\cellcolor{gray!15}\textbf{MESA} & \cellcolor{gray!15}\textbf{31.0} & \cellcolor{gray!15}\textbf{8.6} & \cellcolor{gray!15}75.8 & \cellcolor{gray!15}93.5 \\
\bottomrule
\end{tabular}
\label{tab:loss_perturb}
\end{table}

\begin{figure}[t]
\centering
\begin{subfigure}{0.23\textwidth}
    \centering
    \includegraphics[width=\linewidth]{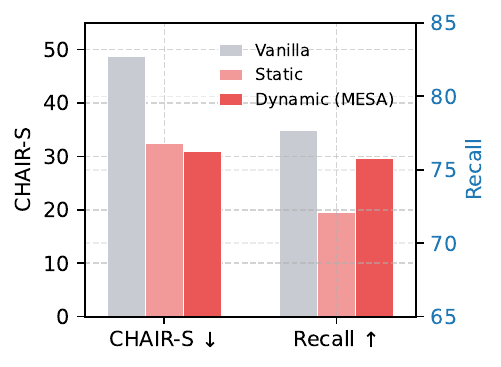}
    \caption{Perturbation weights.}
    \label{fig:perturb_weight}
\end{subfigure}
\hfill
\begin{subfigure}{0.23\textwidth}
    \centering
    \includegraphics[width=\linewidth]{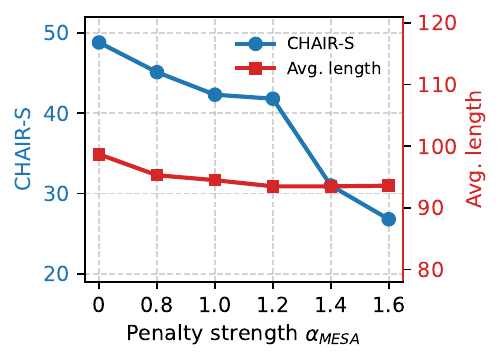}
    \caption{Penalty strength $\alpha_{MESA}$.}
    \label{fig:alpha_mesa}
\end{subfigure}
\caption{Ablation of key parameters. (a) Effect of different perturbation weight combinations in the hallucination-guided loss $\mathcal{L}_{\text{hall}}$ for training perturbation $\delta$. (b) Effect of varying penalty strength $\alpha$ on the steering direction.}
\label{fig:parameter_analysis}
\end{figure}

\subsubsection{Baselines.}
We adopt the widely used LLaVA-v1.5~\cite{liu2024improved} and Qwen-VL~\cite{bai2023qwen} as baseline LVLMs. We further compare against three representative categories of prior methods: (1) decoding-based approaches that mitigate language priors (e.g., VCD~\cite{leng2024vcd}, ICD~\cite{wang2024icd}), (2) attention-based approaches that adjust visual attention weights (e.g., VAF~\cite{yin2025clearsight}, ICT~\cite{chen2025ict}), and (3) latent space steering methods that modulate internal representations (e.g., VTI~\cite{liu2025vti}, Nullu~\cite{yang2025nullu}).

\subsubsection{Implementation Details.}
To evaluate the effectiveness of MESA in mitigating hallucinations, we adopt two representative LVLMs: LLaVA-v1.5~\cite{liu2023visual} and Qwen-VL~\cite{bai2023qwen}.
For perturbation training, we first construct a supervision set $\mathcal{S}$ using multiple visual degradation strategies, including Gaussian noise~\cite{leng2024vcd}, patch masking~\cite{liu2025vti}, blur~\cite{zhang2018blur}, jigsaw shuffling~\cite{kim2002jigsaw}, and a text-only baseline. For each variant, we run the LVLM once and cache the first-step token logits, together with those from the original image, forming prior and reference distributions.
In Stage II, we freeze all model parameters and optimize a visual perturbation $\delta$ applied to vision tokens after the multimodal projector. The perturbation is parameterized by a lightweight two-layer MLP and trained for 10 epochs using AdamW~\cite{loshchilov2017adamw} with a learning rate of $1\mathrm{e}{-3}$. 
The hallucination objective $\mathcal{L}_{\text{hall}}$ is computed over the top-50 tokens, while $\mathcal{L}_{\text{preserve}}$ excludes the highest-probability ones (e.g., top-5) to avoid over-constraining dominant tokens.
To constrain the perturbation, we impose an fidelity bound with $\epsilon = 1$.
During inference, we inject the learned direction into all transformer layers, with a scaling coefficient $\alpha = 1.4$ for CHAIR. For all methods except contrastive decoding (i.e., VCD~\cite{leng2024vcd}, ICD~\cite{wang2024icd}), we adopt greedy decoding for a fair comparison, eliminating stochastic variance and ensuring that performance differences are solely attributable to the intervention strategy. More implementation details are provided in the appendix.

\begin{table*}[t!]
\centering
\caption{An ablation study of different decoding strategies.}
\setlength{\tabcolsep}{4pt}
\renewcommand{\arraystretch}{1.1}
\begin{tabular}{cccccc|cccccc}
\hline
Decoding & Method & $\mathrm{CHAIR}_S$\scriptsize$\downarrow$ 
& $\mathrm{CHAIR}_I$\scriptsize$\downarrow$ 
& Recall\scriptsize$\uparrow$ 
& Avg. len\scriptsize$\uparrow$ 
& Decoding & Method & $\mathrm{CHAIR}_S$\scriptsize$\downarrow$ 
& $\mathrm{CHAIR}_I$\scriptsize$\downarrow$ 
& Recall\scriptsize$\uparrow$ 
& Avg. len\scriptsize$\uparrow$ \\ 
\hline

\multirow{2}{*}{Top P} 
& Vanilla & 52.4 & 15.4 & 76.2 & 98.5 
& \multirow{2}{*}{Top P+Temp.} & Vanilla & 50.6 & 14.4 & 78.3 & 98.2 \\
& \cellcolor{gray!15}\textbf{MESA}    
& \cellcolor{gray!15}36.2 
& \cellcolor{gray!15}8.7  
& \cellcolor{gray!15}75.3 
& \cellcolor{gray!15}90.1 
&                              
& \cellcolor{gray!15}\textbf{MESA}    
& \cellcolor{gray!15}39.6 
& \cellcolor{gray!15}10.1 
& \cellcolor{gray!15}75.2 
& \cellcolor{gray!15}88.8 \\
\hline

\multirow{2}{*}{Top K} 
& Vanilla & 52.8 & 16.0 & 74.5 & 102.9 
& \multirow{2}{*}{Top P+Temp.} & Vanilla & 51.2 & 14.9 & 76.7 & 98.2 \\
& \cellcolor{gray!15}\textbf{MESA}    
& \cellcolor{gray!15}38.0 
& \cellcolor{gray!15}9.9  
& \cellcolor{gray!15}71.5 
& \cellcolor{gray!15}92.3 
&                              
& \cellcolor{gray!15}\textbf{MESA}    
& \cellcolor{gray!15}39.8 
& \cellcolor{gray!15}9.2  
& \cellcolor{gray!15}74.6 
& \cellcolor{gray!15}89.8 \\
\hline

\multirow{2}{*}{Temp.} 
& Vanilla & 50.8 & 15.0 & 77.6 & 99.0 
& \multirow{2}{*}{Greedy} & Vanilla & 48.8 & 13.4 & 77.7 & 98.7 \\
& \cellcolor{gray!15}\textbf{MESA}    
& \cellcolor{gray!15}35.6 
& \cellcolor{gray!15}9.3  
& \cellcolor{gray!15}73.4 
& \cellcolor{gray!15}89.2 
&                              
& \cellcolor{gray!15}\textbf{MESA}    
& \cellcolor{gray!15}31.0 
& \cellcolor{gray!15}8.6  
& \cellcolor{gray!15}75.8 
& \cellcolor{gray!15}93.5 \\
\hline

\end{tabular}
\label{tab:abla_decoding}
\end{table*}

\begin{figure*}[t]
\centering
\begin{subfigure}{0.23\textwidth}
    \centering
    \includegraphics[width=\linewidth]{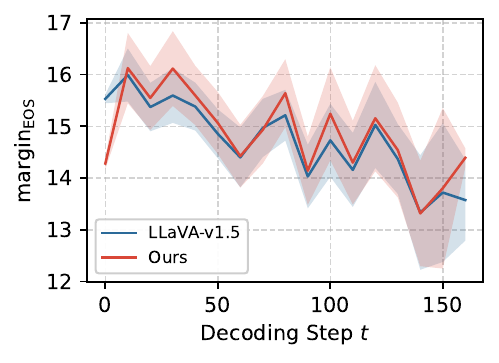}
    \caption{EOS margin (MESA).}
    \label{fig:eos_mesa}
\end{subfigure}
\hfill
\begin{subfigure}{0.23\textwidth}
    \centering
    \includegraphics[width=\linewidth]{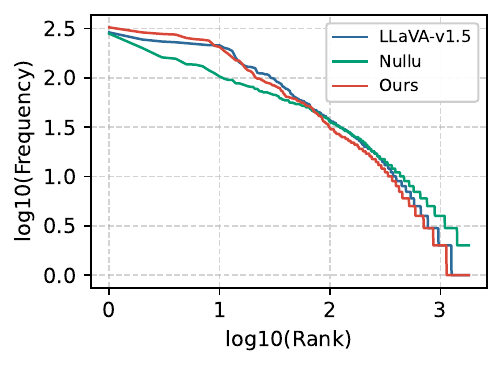}
    \caption{Zipf distribution (MESA).}
    \label{fig:zipf_mesa}
\end{subfigure}
\hfill
\begin{subfigure}{0.23\textwidth}
    \centering
    \includegraphics[width=\linewidth]{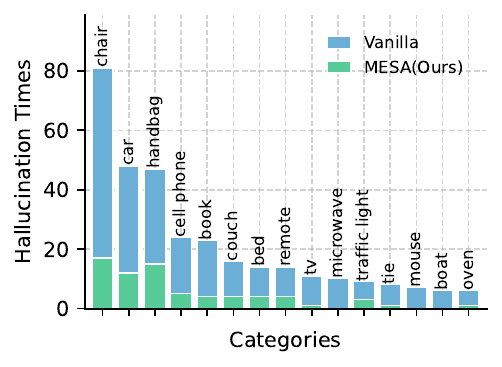}
    \caption{Hallucination Word Reduction (CHAIR). }
    \label{fig:error_remove_chair}
\end{subfigure}
\hfill
\begin{subfigure}{0.23\textwidth}
    \centering
    \includegraphics[width=\linewidth]{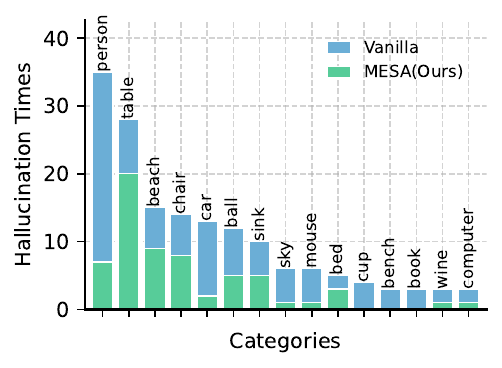}
    \caption{Hallucination Word Reduction (AMBER).}
    \label{fig:error_remove_amber}
\end{subfigure}
\caption{
Further analysis of MESA. 
(a)-(b) Generation behavior on CHAIR with LLaVA-v1.5: MESA preserves the EOS margin and maintains the Zipfian token distribution, indicating minimal disruption to the original generation dynamics. 
(c)-(d) Word-level hallucination reduction on CHAIR and AMBER: MESA consistently suppresses high-frequency hallucinated object words compared to the baseline.
}
\label{fig:generation_behavior_mesa}
\end{figure*}

\subsection{Main Results}

\subsubsection{Results on CHAIR}
We first evaluate our model on open-ended caption generation using the CHAIR benchmark. The evaluation is performed on 1000 randomly selected images from the MSCOCO 2014 validation set~\cite{lin2014microsoft}, using the prompt “Please describe this image in
detail.” 
As shown in ~\autoref{tab:chair}, the proposed MESA consistently outperforms all compared methods in terms of both $\mathrm{CHAIR}_S$ and $\mathrm{CHAIR}_I$. 
Although ICD achieves the highest average output length, its hallucination metrics are the worst, likely because longer outputs increase the chance of generating ungrounded content.
In contrast, MESA achieves the lowest hallucination while maintaining competitive generation length. In contrast to prior approaches, especially latent space steering methods such as VTI and Nullu, our method does not reduce hallucination by producing conservative or shorter outputs. Instead, it preserves output length while effectively mitigating hallucination, which demonstrates the effectiveness of the proposed controlled steering direction.

\subsubsection{Results on POPE}
The results on POPE are shown in ~\autoref{tab:pope}. For the POPE benchmark, we evaluate across three sampling settings: random, popular, and adversarial. Here, we report results on MSCOCO and GQA, which are drawn from different image sources, while the corresponding results on A-OKVQA are provided in the appendix. Under the same decoding strategy (i.e., greedy), the proposed MESA average improves LLaVA-v1.5 on the MSCOCO dataset by 5.6\% in accuracy and 3.12\% in F1 score; for Qwen-VL, it achieves gains of 3.18\% in accuracy and 2\% in F1 score. Notably, compared with latent space steering methods (i.e., Nullu~\cite{yang2025nullu} and VTI~\cite{liu2025vti}), MESA yields 1.21\%–4.86\% improvements in accuracy on the LLaVA-v1.5 backbone, and outperforms them by 0.96\%–18.23\% in F1 score on Qwen-VL. In addition, compared to attention-based methods (i.e., VAF~\cite{yin2025clearsight} and ICT~\cite{chen2025ict}), our method achieves the best results in most cases. Overall, these results consistently validate the robustness and superiority of MESA across different evaluation settings.

\subsubsection{Results on AMBER}
\autoref{tab:amber} reports the comparison results on the AMBER benchmark, which provides a comprehensive evaluation suite spanning both generative and discriminative metrics, including existence, attribute, and relation hallucinations and so on. As shown in \autoref{tab:amber}, our method achieves the highest discriminative Accuracy (84.3) and the lowest generative CHAIR score (6.4).
We further present a breakdown of hallucination types in the discriminative task in ~\autoref{fig:radar_amber}. Compared with other latent-space steering methods, MESA demonstrates superior performance, particularly on relation and existence hallucinations, highlighting its effectiveness in mitigating multimodal hallucinations.

\subsubsection{Results on LLaVA-Bench}
Following ~\cite{yang2025nullu,leng2024vcd}, we leverage LLaVA-Bench, a pairwise evaluation benchmark, to qualitatively assess model performance using GPT-4o-aided evaluation. Specifically, responses from two methods are compared side-by-side. Accuracy measures the factual correctness of the generated responses, while Detailedness evaluates the richness and completeness of the descriptions. The evaluation prompts are provided in the appendix. As shown in ~\autoref{tab:llava_bench}, MESA effectively mitigates hallucinations while preserving or enhancing descriptive quality, demonstrating strong generalization across different intervention strategies.

\subsection{Ablation Study}

\subsubsection{Ablation of training loss}
As shown in \autoref{tab:loss_perturb}, we ablate the two objectives for training the perturbation $\delta$, namely $\mathcal{L}_{\text{hall}}$ and $\mathcal{L}_{\text{preserve}}$, on LLaVA-v1.5 under CHAIR. Introducing $\mathcal{L}_{\text{hall}}$ produces cleaner and more disentangled hallucination signals, leading to a more effective steering direction and reduced hallucinations. Further incorporating $\mathcal{L}_{\text{preserve}}$ (MESA) enhances mitigation while maintaining stable generation behavior.

\subsubsection{Ablation of perturbation designs}
Additionally, to construct the supervision set for training $\delta$ (as illustrated in \autoref{eq:supversion_set}), we further investigate the impact of different perturbation types and weighting strategies, as shown in the lower part of \autoref{tab:loss_perturb}.
First, we compare simple perturbations used in prior work~\cite{leng2024vcd,chen2025ict} (e.g., Gaussian noise) with a more diverse set, including JPEG compression and grayscale transformations. Our method performs better, indicating that modeling diverse perturbation types is crucial. Notably, the jigsaw-grid perturbation disrupts spatial structure, improving robustness to relation hallucinations (see \autoref{fig:radar_amber}).
Second, we introduce a dynamic weighting strategy (\autoref{eq:dynamic_weight}) that emphasizes perturbations causing larger distributional shifts, effectively prioritizing harder hallucination cases and boosting performance.

\subsubsection{Ablation of steering strength $\alpha$}
\autoref{fig:alpha_mesa} illustrates the effect of varying the steering strength $\alpha$ in \autoref{eq:steer} for the direction $d^{(l)}$. As $\alpha$ increases, MESA maintains relatively stable generation length compared to VTI~\cite{liu2025vti}, indicating that the intervention does not overly disrupt the decoding process. Meanwhile, hallucination metrics consistently decrease with larger $\alpha$, demonstrating the effectiveness of stronger steering in suppressing hallucinated content. However, when $\alpha = 1.6$, Recall drops significantly, suggesting that excessive steering may impair content coverage. Therefore, for the CHAIR benchmark, we set $\alpha = 1.4$ as a trade-off between hallucination reduction and semantic completeness.

\subsection{Further Analysis}
\subsubsection{Effect of different decoding strategies.}
Besides greedy decoding, we also present an ablation study  with various decoding strategies on the CHAIR benchmark using LLaVA-v1.5. Following prior work~\cite{an2025agla}, the experiment
includes five kinds of additional decoding strategies: Top P
sampling~\cite{holtzman2019curious} (p = 0.7), Top K sampling~\cite{fan2018hierarchical} (k = 50),
Temperature sampling~\cite{ackley1985learning}(t = 0.5), Top P sampling with
temperature (p = 0.7 and t = 0.5) and Top K sampling with
temperature (k = 50 and t = 0.5). The results in ~\autoref{tab:abla_decoding} show that MESA consistently contributes to mitigating hallucinations, irrespective of the decoding strategies adopted.

\subsubsection{Generation behavior analysis.}
In \autoref{sec:motivation}, we analyze that existing latent space steering methods~\cite{yang2025nullu,liu2025vti} reduce hallucinations at the cost of altering generation behavior, fundamentally affecting the EOS token dynamics and the overall token distribution.
\autoref{fig:generation_behavior_mesa} presents the performance of MESA on these aspects under the CHAIR benchmark using LLaVA-v1.5. For the EOS margin metric (see \autoref{eq:eos}), VTI~\cite{liu2025vti} significantly shifts the distribution, leading to premature termination, whereas our method remains largely consistent with LLaVA-v1.5. Furthermore, from the perspective of token distribution characterized by Zipf’s law~\cite{mikhaylovskiy2025zipf}, our method introduces minimal deviation compared to Nullu~\cite{yang2025nullu}.
These results demonstrate that MESA effectively mitigates hallucinations while preserving the original generation behavior.

\subsubsection{Visualization}
The word-level hallucination statistics in ~\autoref{fig:error_remove_chair} and ~\autoref{fig:error_remove_amber} show that MESA consistently reduces hallucinations across CHAIR and AMBER with LLaVA-v1.5. The reduction is most pronounced for high-frequency categories (e.g., chair, car, person), indicating effective suppression of dominant hallucination sources, while low-frequency tokens remain largely unaffected. These results suggest that MESA mitigates hallucination at the lexical level without distorting the overall generation distribution.


\section{Conclusion}
In this paper, we revisit latent space steering for hallucination mitigation in LVLMs and, for the first time, reveal that existing methods reduce hallucinations at the cost of altering generation behavior due to entangled steering directions. To address this issue, we propose MESA, an effective plug-and-play framework that induces cleaner hallucination-specific representations via controlled perturbations, enabling more selective and disentangled steering. By jointly optimizing hallucination-enhancing and distribution-preserving objectives, MESA effectively and consistently reduces hallucinations while maintaining the model’s original generation dynamics. Extensive experiments across both generative and discriminative benchmarks demonstrate that MESA consistently achieves superior hallucination mitigation and better preserves generation behavior compared to prior methods.





\bibliographystyle{ACM-Reference-Format}
\bibliography{reference}

@article{huynh2025visual,
  title={Visual question answering: from early developments to recent advances--a survey},
  author={Huynh, Ngoc Dung and Bouadjenek, Mohamed Reda and Aryal, Sunil and Razzak, Imran and Hacid, Hakim},
  journal={arXiv preprint arXiv:2501.03939},
  year={2025}
}

@inproceedings{li2024configure,
  title={How to configure good in-context sequence for visual question answering},
  author={Li, Li and Peng, Jiawei and Chen, Huiyi and Gao, Chongyang and Yang, Xu},
  booktitle={Proceedings of the IEEE/CVF Conference on Computer Vision and Pattern Recognition},
  pages={26710--26720},
  year={2024}
}

@article{hartsock2024vision,
  title={Vision-language models for medical report generation and visual question answering: A review},
  author={Hartsock, Iryna and Rasool, Ghulam},
  journal={Frontiers in artificial intelligence},
  volume={7},
  pages={1430984},
  year={2024},
  publisher={Frontiers Media SA}
}

@article{hossain2019comprehensive,
  title={A comprehensive survey of deep learning for image captioning},
  author={Hossain, MD Zakir and Sohel, Ferdous and Shiratuddin, Mohd Fairuz and Laga, Hamid},
  journal={ACM Computing Surveys (CsUR)},
  volume={51},
  number={6},
  pages={1--36},
  year={2019},
  publisher={ACM New York, NY, USA}
}

@article{zhang2024vision,
  title={Vision-language models for vision tasks: A survey},
  author={Zhang, Jingyi and Huang, Jiaxing and Jin, Sheng and Lu, Shijian},
  journal={IEEE transactions on pattern analysis and machine intelligence},
  volume={46},
  number={8},
  pages={5625--5644},
  year={2024},
  publisher={IEEE}
}

@article{chen2024detecting,
  title={Detecting and evaluating medical hallucinations in large vision language models},
  author={Chen, Jiawei and Yang, Dingkang and Wu, Tong and Jiang, Yue and Hou, Xiaolu and Li, Mingcheng and Wang, Shunli and Xiao, Dongling and Li, Ke and Zhang, Lihua},
  journal={arXiv preprint arXiv:2406.10185},
  year={2024}
}

@article{liu2024survey,
  title={A survey on hallucination in large vision-language models},
  author={Liu, Hanchao and Xue, Wenyuan and Chen, Yifei and Chen, Dapeng and Zhao, Xiutian and Wang, Ke and Hou, Liping and Li, Rongjun and Peng, Wei},
  journal={arXiv preprint arXiv:2402.00253},
  year={2024}
}

@article{dong2024benchmarking,
  title={Benchmarking and improving detail image caption},
  author={Dong, Hongyuan and Li, Jiawen and Wu, Bohong and Wang, Jiacong and Zhang, Yuan and Guo, Haoyuan},
  journal={arXiv preprint arXiv:2405.19092},
  year={2024}
}

@article{huang2025shield,
  title={SHIELD: Suppressing Hallucinations In LVLM Encoders via Bias and Vulnerability Defense},
  author={Huang, Yiyang and Shi, Liang and Zhang, Yitian and Xu, Yi and Fu, Yun},
  journal={arXiv preprint arXiv:2510.16596},
  year={2025}
}

@article{li2024surveying,
  title={Surveying the mllm landscape: A meta-review of current surveys},
  author={Li, Ming and Chen, Keyu and Bi, Ziqian and Liu, Ming and Song, Xinyuan and Jiang, Zekun and Wang, Tianyang and Peng, Benji and Niu, Qian and Liu, Junyu and others},
  journal={arXiv preprint arXiv:2409.18991},
  year={2024}
}

@inproceedings{ouali2024clip,
  title={Clip-dpo: Vision-language models as a source of preference for fixing hallucinations in lvlms},
  author={Ouali, Yassine and Bulat, Adrian and Martinez, Brais and Tzimiropoulos, Georgios},
  booktitle={European Conference on Computer Vision},
  pages={395--413},
  year={2024},
  organization={Springer}
}

@article{gilardi2023chatgpt,
  title={ChatGPT outperforms crowd workers for text-annotation tasks},
  author={Gilardi, Fabrizio and Alizadeh, Meysam and Kubli, Ma{\"e}l},
  journal={Proceedings of the National Academy of Sciences},
  volume={120},
  number={30},
  pages={e2305016120},
  year={2023},
  publisher={National Academy of Sciences}
}

@article{touvron2023llama,
  title={Llama: Open and efficient foundation language models},
  author={Touvron, Hugo and Lavril, Thibaut and Izacard, Gautier and Martinet, Xavier and Lachaux, Marie-Anne and Lacroix, Timoth{\'e}e and Rozi{\`e}re, Baptiste and Goyal, Naman and Hambro, Eric and Azhar, Faisal and others},
  journal={arXiv preprint arXiv:2302.13971},
  year={2023}
}

@article{chowdhery2023palm,
  title={Palm: Scaling language modeling with pathways},
  author={Chowdhery, Aakanksha and Narang, Sharan and Devlin, Jacob and Bosma, Maarten and Mishra, Gaurav and Roberts, Adam and Barham, Paul and Chung, Hyung Won and Sutton, Charles and Gehrmann, Sebastian and others},
  journal={Journal of machine learning research},
  volume={24},
  number={240},
  pages={1--113},
  year={2023}
}

@article{chiang2023vicuna,
  title={Vicuna: An open-source chatbot impressing gpt-4 with 90\%* chatgpt quality},
  author={Chiang, Wei-Lin and Li, Zhuohan and Lin, Ziqing and Sheng, Ying and Wu, Zhanghao and Zhang, Hao and Zheng, Lianmin and Zhuang, Siyuan and Zhuang, Yonghao and Gonzalez, Joseph E and others},
  journal={See https://vicuna. lmsys. org (accessed 14 April 2023)},
  volume={2},
  number={3},
  pages={6},
  year={2023}
}

@article{bai2023qwen,
  title={Qwen technical report},
  author={Bai, Jinze and Bai, Shuai and Chu, Yunfei and Cui, Zeyu and Dang, Kai and Deng, Xiaodong and Fan, Yang and Ge, Wenbin and Han, Yu and Huang, Fei and others},
  journal={arXiv preprint arXiv:2309.16609},
  year={2023}
}

@inproceedings{li2022blip,
  title={Blip: Bootstrapping language-image pre-training for unified vision-language understanding and generation},
  author={Li, Junnan and Li, Dongxu and Xiong, Caiming and Hoi, Steven},
  booktitle={International conference on machine learning},
  pages={12888--12900},
  year={2022},
  organization={PMLR}
}

@inproceedings{li2023blip,
  title={Blip-2: Bootstrapping language-image pre-training with frozen image encoders and large language models},
  author={Li, Junnan and Li, Dongxu and Savarese, Silvio and Hoi, Steven},
  booktitle={International conference on machine learning},
  pages={19730--19742},
  year={2023},
  organization={PMLR}
}

@inproceedings{liu2024improved,
  title={Improved baselines with visual instruction tuning},
  author={Liu, Haotian and Li, Chunyuan and Li, Yuheng and Lee, Yong Jae},
  booktitle={Proceedings of the IEEE/CVF conference on computer vision and pattern recognition},
  pages={26296--26306},
  year={2024}
}

@article{liu2023visual,
  title={Visual instruction tuning},
  author={Liu, Haotian and Li, Chunyuan and Wu, Qingyang and Lee, Yong Jae},
  journal={Advances in neural information processing systems},
  volume={36},
  pages={34892--34916},
  year={2023}
}

@article{zhu2023minigpt,
  title={Minigpt-4: Enhancing vision-language understanding with advanced large language models},
  author={Zhu, Deyao and Chen, Jun and Shen, Xiaoqian and Li, Xiang and Elhoseiny, Mohamed},
  journal={arXiv preprint arXiv:2304.10592},
  year={2023}
}

@inproceedings{chen2024internvl,
  title={Internvl: Scaling up vision foundation models and aligning for generic visual-linguistic tasks},
  author={Chen, Zhe and Wu, Jiannan and Wang, Wenhai and Su, Weijie and Chen, Guo and Xing, Sen and Zhong, Muyan and Zhang, Qinglong and Zhu, Xizhou and Lu, Lewei and others},
  booktitle={Proceedings of the IEEE/CVF conference on computer vision and pattern recognition},
  pages={24185--24198},
  year={2024}
}

@article{ye2024mplug,
  title={mplug-owl3: Towards long image-sequence understanding in multi-modal large language models},
  author={Ye, Jiabo and Xu, Haiyang and Liu, Haowei and Hu, Anwen and Yan, Ming and Qian, Qi and Zhang, Ji and Huang, Fei and Zhou, Jingren},
  journal={arXiv preprint arXiv:2408.04840},
  year={2024}
}

@article{bai2024hallucination,
  title={Hallucination of multimodal large language models: A survey},
  author={Bai, Zechen and Wang, Pichao and Xiao, Tianjun and He, Tong and Han, Zongbo and Zhang, Zheng and Shou, Mike Zheng},
  journal={arXiv preprint arXiv:2404.18930},
  year={2024}
}

@inproceedings{chen2025ict,
  title={Ict: Image-object cross-level trusted intervention for mitigating object hallucination in large vision-language models},
  author={Chen, Junzhe and Zhang, Tianshu and Huang, Shiyu and Niu, Yuwei and Zhang, Linfeng and Wen, Lijie and Hu, Xuming},
  booktitle={Proceedings of the Computer Vision and Pattern Recognition Conference},
  pages={4209--4221},
  year={2025}
}

@inproceedings{su2025activation,
  title={Activation steering decoding: Mitigating hallucination in large vision-language models through bidirectional hidden state intervention},
  author={Su, Jingran and Chen, Jingfan and Li, Hongxin and Chen, Yuntao and Qing, Li and Zhang, Zhaoxiang},
  booktitle={Proceedings of the 63rd Annual Meeting of the Association for Computational Linguistics (Volume 1: Long Papers)},
  pages={12964--12974},
  year={2025}
}

@article{liu2023mitigating,
  title={Mitigating hallucination in large multi-modal models via robust instruction tuning},
  author={Liu, Fuxiao and Lin, Kevin and Li, Linjie and Wang, Jianfeng and Yacoob, Yaser and Wang, Lijuan},
  journal={arXiv preprint arXiv:2306.14565},
  year={2023}
}

@inproceedings{yu2024rlhf,
  title={Rlhf-v: Towards trustworthy mllms via behavior alignment from fine-grained correctional human feedback},
  author={Yu, Tianyu and Yao, Yuan and Zhang, Haoye and He, Taiwen and Han, Yifeng and Cui, Ganqu and Hu, Jinyi and Liu, Zhiyuan and Zheng, Hai-Tao and Sun, Maosong and others},
  booktitle={Proceedings of the IEEE/CVF Conference on Computer Vision and Pattern Recognition},
  pages={13807--13816},
  year={2024}
}

@inproceedings{yu2025rlaif,
  title={Rlaif-v: Open-source ai feedback leads to super gpt-4v trustworthiness},
  author={Yu, Tianyu and Zhang, Haoye and Li, Qiming and Xu, Qixin and Yao, Yuan and Chen, Da and Lu, Xiaoman and Cui, Ganqu and Dang, Yunkai and He, Taiwen and others},
  booktitle={Proceedings of the Computer Vision and Pattern Recognition Conference},
  pages={19985--19995},
  year={2025}
}

@inproceedings{leng2024vcd,
  title={Mitigating object hallucinations in large vision-language models through visual contrastive decoding},
  author={Leng, Sicong and Zhang, Hang and Chen, Guanzheng and Li, Xin and Lu, Shijian and Miao, Chunyan and Bing, Lidong},
  booktitle={Proceedings of the IEEE/CVF Conference on Computer Vision and Pattern Recognition},
  pages={13872--13882},
  year={2024}
}

@inproceedings{wang2024icd,
  title={Mitigating hallucinations in large vision-language models with instruction contrastive decoding},
  author={Wang, Xintong and Pan, Jingheng and Ding, Liang and Biemann, Chris},
  booktitle={Findings of the Association for Computational Linguistics: ACL 2024},
  pages={15840--15853},
  year={2024}
}

@inproceedings{yin2025clearsight,
  title={Clearsight: Visual signal enhancement for object hallucination mitigation in multimodal large language models},
  author={Yin, Hao and Si, Guangzong and Wang, Zilei},
  booktitle={Proceedings of the Computer Vision and Pattern Recognition Conference},
  pages={14625--14634},
  year={2025}
}

@article{yin2026dynamic,
  title={Dynamic Multimodal Activation Steering for Hallucination Mitigation in Large Vision-Language Models},
  author={Yin, Jianghao and Chen, Qin and Chen, Kedi and Zhou, Jie and Wu, Xingjiao and He, Liang},
  journal={arXiv preprint arXiv:2602.21704},
  year={2026}
}

@inproceedings{yang2025nullu,
  title={Nullu: Mitigating object hallucinations in large vision-language models via halluspace projection},
  author={Yang, Le and Zheng, Ziwei and Chen, Boxu and Zhao, Zhengyu and Lin, Chenhao and Shen, Chao},
  booktitle={Proceedings of the Computer Vision and Pattern Recognition Conference},
  pages={14635--14645},
  year={2025}
}

@inproceedings{liu2025vti,
  title={Reducing hallucinations in large vision-language models via latent space steering},
  author={Liu, Sheng and Ye, Haotian and Zou, James},
  booktitle={The Thirteenth International Conference on Learning Representations},
  year={2025}
}

@inproceedings{an2025agla,
  title={Mitigating object hallucinations in large vision-language models with assembly of global and local attention},
  author={An, Wenbin and Tian, Feng and Leng, Sicong and Nie, Jiahao and Lin, Haonan and Wang, QianYing and Chen, Ping and Zhang, Xiaoqin and Lu, Shijian},
  booktitle={Proceedings of the Computer Vision and Pattern Recognition Conference},
  pages={29915--29926},
  year={2025}
}

@inproceedings{rohrbach2018object,
  title={Object hallucination in image captioning},
  author={Rohrbach, Anna and Hendricks, Lisa Anne and Burns, Kaylee and Darrell, Trevor and Saenko, Kate},
  booktitle={Proceedings of the 2018 Conference on Empirical Methods in Natural Language Processing},
  pages={4035--4045},
  year={2018}
}

@inproceedings{lin2014microsoft,
  title={Microsoft coco: Common objects in context},
  author={Lin, Tsung-Yi and Maire, Michael and Belongie, Serge and Hays, James and Perona, Pietro and Ramanan, Deva and Doll{\'a}r, Piotr and Zitnick, C Lawrence},
  booktitle={European conference on computer vision},
  pages={740--755},
  year={2014},
  organization={Springer}
}

@inproceedings{li2023evaluating,
  title={Evaluating object hallucination in large vision-language models},
  author={Li, Yifan and Du, Yifan and Zhou, Kun and Wang, Jinpeng and Zhao, Wayne Xin and Wen, Ji-Rong},
  booktitle={Proceedings of the 2023 conference on empirical methods in natural language processing},
  pages={292--305},
  year={2023}
}

@inproceedings{schwenk2022okvqa,
  title={A-okvqa: A benchmark for visual question answering using world knowledge},
  author={Schwenk, Dustin and Khandelwal, Apoorv and Clark, Christopher and Marino, Kenneth and Mottaghi, Roozbeh},
  booktitle={European conference on computer vision},
  pages={146--162},
  year={2022},
  organization={Springer}
}

@inproceedings{hudson2019gqa,
  title={Gqa: A new dataset for real-world visual reasoning and compositional question answering},
  author={Hudson, Drew A and Manning, Christopher D},
  booktitle={Proceedings of the IEEE/CVF conference on computer vision and pattern recognition},
  pages={6700--6709},
  year={2019}
}

@article{wang2023amber,
  title={Amber: An llm-free multi-dimensional benchmark for mllms hallucination evaluation},
  author={Wang, Junyang and Wang, Yuhang and Xu, Guohai and Zhang, Jing and Gu, Yukai and Jia, Haitao and Wang, Jiaqi and Xu, Haiyang and Yan, Ming and Zhang, Ji and others},
  journal={arXiv preprint arXiv:2311.07397},
  year={2023}
}

@article{mikhaylovskiy2025zipf,
  title={Zipf’s and Heaps’ Laws for Tokens and LLM-generated Texts},
  author={Mikhaylovskiy, Nikolay},
  journal={Findings of the Association for Computational Linguistics: EMNLP 2025},
  pages={15469--15481},
  year={2025}
}

@article{holtzman2019curious,
  title={The curious case of neural text degeneration},
  author={Holtzman, Ari and Buys, Jan and Du, Li and Forbes, Maxwell and Choi, Yejin},
  journal={arXiv preprint arXiv:1904.09751},
  year={2019}
}

@inproceedings{fan2018hierarchical,
  title={Hierarchical neural story generation},
  author={Fan, Angela and Lewis, Mike and Dauphin, Yann},
  booktitle={Proceedings of the 56th Annual Meeting of the Association for Computational Linguistics (Volume 1: Long Papers)},
  pages={889--898},
  year={2018}
}

@article{ackley1985learning,
  title={A learning algorithm for Boltzmann machines},
  author={Ackley, David H and Hinton, Geoffrey E and Sejnowski, Terrence J},
  journal={Cognitive science},
  volume={9},
  number={1},
  pages={147--169},
  year={1985},
  publisher={Elsevier}
}

@inproceedings{zhang2018blur,
  title={Learning to understand image blur},
  author={Zhang, Shanghang and Shen, Xiaohui and Lin, Zhe and M{\v{e}}ch, Radom{\'\i}r and Costeira, Joao P and Moura, Jos{\'e} MF},
  booktitle={Proceedings of the IEEE conference on computer vision and pattern recognition},
  pages={6586--6595},
  year={2018}
}

@article{kim2002jigsaw,
  title={Jigsaw image mosaics},
  author={Kim, Junhwan and Pellacini, Fabio and others},
  journal={ACM Transactions on Graphics},
  volume={21},
  number={3},
  pages={657--664},
  year={2002}
}

@article{loshchilov2017adamw,
  title={Decoupled weight decay regularization},
  author={Loshchilov, Ilya and Hutter, Frank},
  journal={arXiv preprint arXiv:1711.05101},
  year={2017}
}

@article{hoscilowicz2025adversarial,
  title={Adversarial Confusion Attack: Disrupting Multimodal Large Language Models},
  author={Hoscilowicz, Jakub and Janicki, Artur},
  journal={arXiv preprint arXiv:2511.20494},
  year={2025}
}

@article{mackiewicz1993principal,
  title={Principal components analysis (PCA)},
  author={Ma{\'c}kiewicz, Andrzej and Ratajczak, Waldemar},
  journal={Computers \& Geosciences},
  volume={19},
  number={3},
  pages={303--342},
  year={1993},
  publisher={Elsevier}
}

@article{ji2023survey,
  title={Survey of hallucination in natural language generation},
  author={Ji, Ziwei and Lee, Nayeon and Frieske, Rita and Yu, Tiezheng and Su, Dan and Xu, Yan and Ishii, Etsuko and Bang, Ye Jin and Madotto, Andrea and Fung, Pascale},
  journal={ACM computing surveys},
  volume={55},
  number={12},
  pages={1--38},
  year={2023},
  publisher={ACM New York, NY}
}

\appendix

\cleardoublepage
\section{Detailed Experimental Setup}

\subsection{Metrics for CHAIR Benchmark}
In this paper, we have reported  $\mathrm{CHAIR}_S$ and $\mathrm{CHAIR}_I$ as the evaluation metrics. The calculation can be defined as follows:
\begin{equation}
\begin{aligned}
\mathrm{CHAIR}_S &= \frac{\left| \text{sentences with hallucinated objects} \right|}
           {\left| \text{all sentences} \right|}, \\
\mathrm{CHAIR}_I &= \frac{\left| \text{hallucinated objects} \right|}
           {\left| \text{all objects mentioned} \right|}.
\end{aligned}
\end{equation}
The evaluation primarily consists of two complementary dimensions: a sentence-level metric, $\mathrm{CHAIR}_S$, and an instance-level metric, $\mathrm{CHAIR}_I$. In this paper, we conduct evaluation on 1,000 randomly sampled images from the MSCOCO 2014 validation set, using the prompt \textit{"Please describe this image in detail."} to elicit descriptive captions. To ensure a fair comparison across methods, all generated captions are truncated to a maximum length of 512 tokens. This standardized setup mitigates variability arising from generation length and enables a consistent assessment of hallucination behaviors across models.

\subsection{GPT-4o Assisted Evaluation for LLaVA-Bench}
Following prior work (e.g., AGLA, Nullu), we evaluate the overall quality of generated captions on the LLaVA-Bench dataset using GPT-4o as an automatic judge. Specifically, GPT-4o assigns scores to each response along two dimensions—correctness and detailedness—on a scale from 0 to 10. A higher correctness score indicates fewer hallucinations and better alignment with the visual content, while detailedness reflects the richness and completeness of the description. The evaluation explicitly penalizes mentions of objects that are absent from the image, as well as inaccuracies in attributes, colors, spatial positions, and object relationships. This protocol enables a fine-grained assessment of both factual reliability and descriptive quality, providing a balanced view of model performance. The full evaluation prompt is provided in~\autoref{tab:gpt4o_prompt}.

\begin{table*}[t]
\centering
\caption{The prompt used for GPT-4o-aided evaluation.}
\label{tab:gpt4o_prompt}
\setlength{\tabcolsep}{6pt}
\renewcommand{\arraystretch}{0.9}
\begin{tabular}{p{0.95\textwidth}}
\hline\hline
\textbf{GPT-4o Prompt} \\
\hline

\begin{minipage}{\linewidth}
\begin{Verbatim}[breaklines=true, breakanywhere=true, fontsize=\small]
You are an AI designed to evaluate and score the performance of two AI assistants in describing a given image. Your primary focus is on the accuracy and detailedness of their descriptions. You will assess the accuracy by checking for hallucinations - any part of the description that is inconsistent with the image content. For detailedness, you will consider how rich the response is in necessary details, excluding any hallucinated parts. You will provide scores on a scale from 1 to 10 for each assistant separately, based on these criteria. After scoring, you will offer an explanation for your evaluation, ensuring it is free from bias and not influenced by the order of presentation of the responses.

Input format:
[Assistant 1] {Response 1} [End of Assistant 1]
[Assistant 2] {Response 2} [End of Assistant 2]

Output format:
Accuracy: 
Scores of the two answers: 
Reason:

Detailedness: 
Scores of the two answers: 
Reason:
\end{Verbatim}
\end{minipage}

\\
\hline\hline
\end{tabular}
\label{tab:prompt_gpt4o}
\end{table*}

\begin{table*}[t!]
\centering
\caption{Performance Comparison under Different Settings on MSCOCO with POPE Evaluation}
\setlength{\tabcolsep}{4pt}
\renewcommand{\arraystretch}{0.9}

\begin{tabular}{l|l|cccc|cccc}
\toprule
\multirow{2}{*}{Setting} & \multirow{2}{*}{Method} 
& \multicolumn{4}{c|}{\textbf{LLaVA-v1.5}} 
& \multicolumn{4}{c}{\textbf{Qwen-VL}} \\
\cmidrule(lr){3-6} \cmidrule(lr){7-10}
& 
& Accuracy$\uparrow$ & Precision$\uparrow$ & Recall$\uparrow$ & F1 Score$\uparrow$
& Accuracy$\uparrow$ & Precision$\uparrow$ & Recall$\uparrow$ & F1 Score$\uparrow$ \\
\midrule

\multirow[c]{8}{*}{Random}
& Vanilla & 85.53 & 86.77 & 90.93 & 88.80 & 87.23 & 91.70 & 80.06 & 85.50 \\
& VCD     & 87.73 {\scriptsize \color{lightblue}(↑2.20)} & 91.42 {\scriptsize \color{lightblue}(↑4.65)} & 83.28 {\scriptsize \color{lightblue}(↓7.65)} & 87.16 {\scriptsize \color{lightblue}(↓1.64)} & 88.63 {\scriptsize \color{lightblue}(↑1.40)} & 94.64 {\scriptsize \color{lightblue}(↑2.94)} & 81.91 {\scriptsize \color{lightblue}(↑1.85)} & \textbf{87.81} {\scriptsize \color{lightblue}(↑2.31)} \\
& ICD     & 85.23 {\scriptsize \color{lightblue}(↓0.30)} & 94.44 {\scriptsize \color{lightblue}(↑7.67)} & 74.86 {\scriptsize \color{lightblue}(↓16.07)} & 83.52 {\scriptsize \color{lightblue}(↓5.28)} & 82.43 {\scriptsize \color{lightblue}(↓4.80)} & 98.80 {\scriptsize \color{lightblue}(↑7.10)} & 65.67 {\scriptsize \color{lightblue}(↓14.39)} & 78.89 {\scriptsize \color{lightblue}(↓6.61)} \\
& VAF     & 87.36 {\scriptsize \color{lightblue}(↑1.83)} & 96.82 {\scriptsize \color{lightblue}(↑10.05)} & 77.26 {\scriptsize \color{lightblue}(↓13.67)} & 85.94 {\scriptsize \color{lightblue}(↓2.86)} & 86.80 {\scriptsize \color{lightblue}(↓0.43)} & 98.50 {\scriptsize \color{lightblue}(↑6.80)} & 74.73 {\scriptsize \color{lightblue}(↓5.33)} & 84.98 {\scriptsize \color{lightblue}(↓0.52)} \\
& ICT     & 88.96 {\scriptsize \color{lightblue}(↑3.43)} & 96.87 {\scriptsize \color{lightblue}(↑10.10)} & 80.40 {\scriptsize \color{lightblue}(↓10.53)} & 87.85 {\scriptsize \color{lightblue}(↓0.95)} & 88.06 {\scriptsize \color{lightblue}(↑0.83)} & 96.90 {\scriptsize \color{lightblue}(↑5.20)} & 78.11 {\scriptsize \color{lightblue}(↓1.95)} & 86.55 {\scriptsize \color{lightblue}(↑1.05)} \\
& VTI     & 87.70 {\scriptsize \color{lightblue}(↑2.17)} & 96.24 {\scriptsize \color{lightblue}(↑9.47)} & 78.47 {\scriptsize \color{lightblue}(↓12.46)} & 86.45 {\scriptsize \color{lightblue}(↓2.35)} & 87.97 {\scriptsize \color{lightblue}(↑0.74)} & 96.19 {\scriptsize \color{lightblue}(↑4.49)} & 79.07 {\scriptsize \color{lightblue}(↓0.99)} & 86.79 {\scriptsize \color{lightblue}(↑1.29)} \\
& Nullu   & 89.03 {\scriptsize \color{lightblue}(↑3.50)} & 92.83 {\scriptsize \color{lightblue}(↑6.06)} & 84.60 {\scriptsize \color{lightblue}(↓6.33)} & 88.52 {\scriptsize \color{lightblue}(↓0.28)} & 88.26 {\scriptsize \color{lightblue}(↑1.03)} & 91.47 {\scriptsize \color{lightblue}(↓0.23)} & \textbf{81.40} {\scriptsize \color{lightblue}(↑1.34)} & 86.13 {\scriptsize \color{lightblue}(↑0.63)} \\
& MESA    & \textbf{90.27} {\scriptsize \color{lightblue}(↑4.74)} & 95.81 {\scriptsize \color{lightblue}(↑9.04)} & 85.20 {\scriptsize \color{lightblue}(↓5.73)} & \textbf{90.20} {\scriptsize \color{lightblue}(↑1.40)} & \textbf{88.83} {\scriptsize \color{lightblue}(↑1.60)} & 97.29 {\scriptsize \color{lightblue}(↑5.59)} & 78.87 {\scriptsize \color{lightblue}(↓1.19)} & 87.11 {\scriptsize \color{lightblue}(↑1.61)} \\

\midrule

\multirow[c]{8}{*}{Popular}
& Vanilla & 83.20 & 79.15 & 90.13 & 84.28 & 85.10 & 86.63 & 83.00 & 84.78 \\
& VCD     & 85.38 {\scriptsize \color{lightblue}(↑2.18)} & 86.92 {\scriptsize \color{lightblue}(↑7.77)} & 83.28 {\scriptsize \color{lightblue}(↓6.85)} & 85.06 {\scriptsize \color{lightblue}(↑0.78)} & 87.12 {\scriptsize \color{lightblue}(↑2.02)} & 91.49 {\scriptsize \color{lightblue}(↑4.86)} & 81.85 {\scriptsize \color{lightblue}(↓1.15)} & 86.40 {\scriptsize \color{lightblue}(↑1.62)} \\
& ICD     & 83.73 {\scriptsize \color{lightblue}(↑0.53)} & 91.74 {\scriptsize \color{lightblue}(↑12.59)} & 74.13 {\scriptsize \color{lightblue}(↓16.00)} & 82.00 {\scriptsize \color{lightblue}(↓2.28)} & 81.90 {\scriptsize \color{lightblue}(↓3.20)} & 98.28 {\scriptsize \color{lightblue}(↑11.65)} & 64.93 {\scriptsize \color{lightblue}(↓18.07)} & 78.20 {\scriptsize \color{lightblue}(↓6.58)} \\
& VAF     & 86.46 {\scriptsize \color{lightblue}(↑3.26)} & 94.68 {\scriptsize \color{lightblue}(↑15.53)} & 77.26 {\scriptsize \color{lightblue}(↓12.87)} & 85.09 {\scriptsize \color{lightblue}(↑0.81)} & 85.63 {\scriptsize \color{lightblue}(↑0.53)} & 95.56 {\scriptsize \color{lightblue}(↑8.93)} & 74.73 {\scriptsize \color{lightblue}(↓8.27)} & 83.87 {\scriptsize \color{lightblue}(↓0.91)} \\
& ICT     & 85.86 {\scriptsize \color{lightblue}(↑2.66)} & 94.38 {\scriptsize \color{lightblue}(↑15.23)} & 76.26 {\scriptsize \color{lightblue}(↓13.87)} & 84.36 {\scriptsize \color{lightblue}(↑0.08)} & 87.03 {\scriptsize \color{lightblue}(↑1.93)} & 94.08 {\scriptsize \color{lightblue}(↑7.45)} & 79.46 {\scriptsize \color{lightblue}(↓3.54)} & 86.15 {\scriptsize \color{lightblue}(↑1.37)} \\
& VTI     & 86.37 {\scriptsize \color{lightblue}(↑3.17)} & 93.19 {\scriptsize \color{lightblue}(↑14.04)} & 78.47 {\scriptsize \color{lightblue}(↓11.66)} & 85.20 {\scriptsize \color{lightblue}(↑0.92)} & 86.37 {\scriptsize \color{lightblue}(↑1.27)} & 92.58 {\scriptsize \color{lightblue}(↑5.95)} & 79.07 {\scriptsize \color{lightblue}(↓3.93)} & 85.29 {\scriptsize \color{lightblue}(↑0.51)} \\
& Nullu   & 86.06 {\scriptsize \color{lightblue}(↑2.86)} & 87.00 {\scriptsize \color{lightblue}(↑7.85)} & 84.80 {\scriptsize \color{lightblue}(↓5.33)} & 85.88 {\scriptsize \color{lightblue}(↑1.60)} & 85.43 {\scriptsize \color{lightblue}(↑0.33)} & 86.52 {\scriptsize \color{lightblue}(↓0.11)} & 83.93 {\scriptsize \color{lightblue}(↑0.93)} & 85.21 {\scriptsize \color{lightblue}(↑0.43)} \\
& MESA    & \textbf{87.63} {\scriptsize \color{lightblue}(↑4.43)} & 89.56 {\scriptsize \color{lightblue}(↑10.41)} & 85.20 {\scriptsize \color{lightblue}(↓4.93)} & \textbf{87.32} {\scriptsize \color{lightblue}(↑3.04)} & \textbf{87.20} {\scriptsize \color{lightblue}(↑2.10)} & 94.41 {\scriptsize \color{lightblue}(↑7.78)} & 79.87 {\scriptsize \color{lightblue}(↓3.13)} & \textbf{86.51} {\scriptsize \color{lightblue}(↑1.73)} \\

\midrule

\multirow[c]{8}{*}{Adversarial}
& Vanilla & 75.60 & 70.90 & 89.53 & 78.63 & 78.63 & 78.92 & 84.13 & 81.44 \\
& VCD     & 80.88 {\scriptsize \color{lightblue}(↑5.28)} & 79.45 {\scriptsize \color{lightblue}(↑8.55)} & 83.29 {\scriptsize \color{lightblue}(↓6.24)} & 81.33 {\scriptsize \color{lightblue}(↑2.70)} & 84.26 {\scriptsize \color{lightblue}(↑5.63)} & 85.84 {\scriptsize \color{lightblue}(↑6.92)} & 82.05 {\scriptsize \color{lightblue}(↓2.08)} & 83.90 {\scriptsize \color{lightblue}(↑2.46)} \\
& ICD     & 81.46 {\scriptsize \color{lightblue}(↑5.86)} & 85.59 {\scriptsize \color{lightblue}(↑14.69)} & 75.66 {\scriptsize \color{lightblue}(↓13.87)} & 80.32 {\scriptsize \color{lightblue}(↑1.69)} & 80.80 {\scriptsize \color{lightblue}(↑2.17)} & 94.85 {\scriptsize \color{lightblue}(↑15.93)} & 65.13 {\scriptsize \color{lightblue}(↓19.00)} & 77.23 {\scriptsize \color{lightblue}(↓4.21)} \\
& VAF     & \textbf{83.86} {\scriptsize \color{lightblue}(↑8.26)} & 89.07 {\scriptsize \color{lightblue}(↑18.17)} & 77.20 {\scriptsize \color{lightblue}(↓12.33)} & 82.71 {\scriptsize \color{lightblue}(↑4.08)} & 83.70 {\scriptsize \color{lightblue}(↑5.07)} & 91.06 {\scriptsize \color{lightblue}(↑12.14)} & 74.73 {\scriptsize \color{lightblue}(↓9.40)} & 82.09 {\scriptsize \color{lightblue}(↑0.65)} \\
& ICT     & 83.83 {\scriptsize \color{lightblue}(↑8.23)} & 90.11 {\scriptsize \color{lightblue}(↑19.21)} & 76.00 {\scriptsize \color{lightblue}(↓13.53)} & 82.45 {\scriptsize \color{lightblue}(↑3.82)} & 84.30 {\scriptsize \color{lightblue}(↑5.67)} & 87.91 {\scriptsize \color{lightblue}(↑8.99)} & 79.53 {\scriptsize \color{lightblue}(↓4.60)} & 83.51 {\scriptsize \color{lightblue}(↑2.07)} \\
& VTI     & 83.43 {\scriptsize \color{lightblue}(↑7.83)} & 87.23 {\scriptsize \color{lightblue}(↑16.33)} & 78.33 {\scriptsize \color{lightblue}(↓11.20)} & 82.54 {\scriptsize \color{lightblue}(↑3.91)} & 83.53 {\scriptsize \color{lightblue}(↑4.90)} & 86.88 {\scriptsize \color{lightblue}(↑7.96)} & 79.00 {\scriptsize \color{lightblue}(↓5.13)} & 82.75 {\scriptsize \color{lightblue}(↑1.31)} \\
& Nullu   & 78.66 {\scriptsize \color{lightblue}(↑3.06)} & 76.31 {\scriptsize \color{lightblue}(↑5.41)} & 83.13 {\scriptsize \color{lightblue}(↓6.40)} & 79.57 {\scriptsize \color{lightblue}(↑0.94)} & 84.11 {\scriptsize \color{lightblue}(↑5.48)} & 86.63 {\scriptsize \color{lightblue}(↑7.71)} & 78.00 {\scriptsize \color{lightblue}(↓6.13)} & 82.09 {\scriptsize \color{lightblue}(↑0.65)} \\
& MESA    & 83.23 {\scriptsize \color{lightblue}(↑7.63)} & 81.98 {\scriptsize \color{lightblue}(↑11.08)} & 85.20 {\scriptsize \color{lightblue}(↓4.33)} & \textbf{83.56} {\scriptsize \color{lightblue}(↑4.93)} & \textbf{84.47} {\scriptsize \color{lightblue}(↑5.84)} & 88.81 {\scriptsize \color{lightblue}(↑9.89)} & 79.87 {\scriptsize \color{lightblue}(↓4.26)} & \textbf{84.10} {\scriptsize \color{lightblue}(↑2.66)} \\

\bottomrule
\end{tabular}
\label{tab:pope_mscoco}
\end{table*}

\begin{table*}[t!]
\centering
\caption{Performance Comparison under Different Settings on A-OKVQA with POPE Evaluation}
\setlength{\tabcolsep}{4pt}
\renewcommand{\arraystretch}{0.9}

\begin{tabular}{l|l|cccc|cccc}
\toprule
\multirow{2}{*}{Setting} & \multirow{2}{*}{Method} 
& \multicolumn{4}{c|}{\textbf{LLaVA-v1.5}} 
& \multicolumn{4}{c}{\textbf{Qwen-VL}} \\
\cmidrule(lr){3-6} \cmidrule(lr){7-10}
& 
& Accuracy$\uparrow$ & Precision$\uparrow$ & Recall$\uparrow$ & F1 Score$\uparrow$
& Accuracy$\uparrow$ & Precision$\uparrow$ & Recall$\uparrow$ & F1 Score$\uparrow$ \\
\midrule

\multirow[c]{8}{*}{Random}
& Vanilla & 85.40 & 80.83 & 91.30 & 85.76 & 87.46 & 91.32 & 82.80 & 86.85 \\
& VCD     & 86.15 {\scriptsize \color{lightblue}(↑0.75)} & 85.18 {\scriptsize \color{lightblue}(↑4.35)} & 87.53 {\scriptsize \color{lightblue}(↓3.77)} & 86.34 {\scriptsize \color{lightblue}(↑0.58)} & 89.22 {\scriptsize \color{lightblue}(↑1.76)} & 90.77 {\scriptsize \color{lightblue}(↓0.55)} & 87.32 {\scriptsize \color{lightblue}(↑4.52)} & 89.01 {\scriptsize \color{lightblue}(↑2.16)} \\
& ICD     & 86.53 {\scriptsize \color{lightblue}(↑1.13)} & 90.59 {\scriptsize \color{lightblue}(↑9.76)} & 81.53 {\scriptsize \color{lightblue}(↓9.77)} & 85.82 {\scriptsize \color{lightblue}(↑0.06)} & 82.80 {\scriptsize \color{lightblue}(↓4.66)} & 97.04 {\scriptsize \color{lightblue}(↑5.72)} & 67.67 {\scriptsize \color{lightblue}(↓15.13)} & 79.73 {\scriptsize \color{lightblue}(↓7.12)} \\
& VAF     & 88.73 {\scriptsize \color{lightblue}(↑3.33)} & 91.91 {\scriptsize \color{lightblue}(↑11.08)} & 84.93 {\scriptsize \color{lightblue}(↓6.37)} & 88.28 {\scriptsize \color{lightblue}(↑2.52)} & 87.06 {\scriptsize \color{lightblue}(↓0.40)} & 93.78 {\scriptsize \color{lightblue}(↑2.46)} & 80.73 {\scriptsize \color{lightblue}(↓2.07)} & 86.76 {\scriptsize \color{lightblue}(↓0.09)} \\
& ICT     & 88.96 {\scriptsize \color{lightblue}(↑3.56)} & 92.88 {\scriptsize \color{lightblue}(↑12.05)} & 84.40 {\scriptsize \color{lightblue}(↓6.90)} & 88.43 {\scriptsize \color{lightblue}(↑2.67)} & 89.36 {\scriptsize \color{lightblue}(↑1.90)} & 92.88 {\scriptsize \color{lightblue}(↑1.56)} & 85.80 {\scriptsize \color{lightblue}(↑3.00)} & 89.20 {\scriptsize \color{lightblue}(↑2.35)} \\
& VTI     & 88.13 {\scriptsize \color{lightblue}(↑2.73)} & 90.97 {\scriptsize \color{lightblue}(↑10.14)} & 84.67 {\scriptsize \color{lightblue}(↓6.63)} & 87.71 {\scriptsize \color{lightblue}(↑1.95)} & 87.30 {\scriptsize \color{lightblue}(↓0.16)} & 91.60 {\scriptsize \color{lightblue}(↑0.28)} & 82.13 {\scriptsize \color{lightblue}(↓0.67)} & 86.61 {\scriptsize \color{lightblue}(↓0.24)} \\
& Nullu   & 87.66 {\scriptsize \color{lightblue}(↑2.26)} & 86.78 {\scriptsize \color{lightblue}(↑5.95)} & 88.86 {\scriptsize \color{lightblue}(↓2.44)} & 87.81 {\scriptsize \color{lightblue}(↑2.05)} & 87.53 {\scriptsize \color{lightblue}(↑0.07)} & 91.39 {\scriptsize \color{lightblue}(↑0.07)} & 82.86 {\scriptsize \color{lightblue}(↑0.06)} & 86.92 {\scriptsize \color{lightblue}(↑0.07)} \\
& MESA    & \textbf{89.63} {\scriptsize \color{lightblue}(↑4.23)} & 92.99 {\scriptsize \color{lightblue}(↑12.16)} & 85.73 {\scriptsize \color{lightblue}(↓5.57)} & \textbf{89.21} {\scriptsize \color{lightblue}(↑3.45)} & \textbf{90.03} {\scriptsize \color{lightblue}(↑2.57)} & 93.61 {\scriptsize \color{lightblue}(↑2.29)} & 85.93 {\scriptsize \color{lightblue}(↑3.13)} & \textbf{89.61} {\scriptsize \color{lightblue}(↑2.76)} \\

\midrule

\multirow[c]{8}{*}{Popular}
& Vanilla & 78.10 & 71.71 & 92.80 & 80.90 & 87.36 & 91.06 & 82.86 & 86.77 \\
& VCD     & 81.85 {\scriptsize \color{lightblue}(↑3.75)} & 78.60 {\scriptsize \color{lightblue}(↑6.89)} & 87.53 {\scriptsize \color{lightblue}(↓5.27)} & 82.82 {\scriptsize \color{lightblue}(↑1.92)} & 87.85 {\scriptsize \color{lightblue}(↑0.49)} & 88.10 {\scriptsize \color{lightblue}(↓2.96)} & 87.53 {\scriptsize \color{lightblue}(↑4.67)} & 87.81 {\scriptsize \color{lightblue}(↑1.04)} \\
& ICD     & 81.53 {\scriptsize \color{lightblue}(↑3.43)} & 81.87 {\scriptsize \color{lightblue}(↑10.16)} & 81.00 {\scriptsize \color{lightblue}(↓11.80)} & 81.43 {\scriptsize \color{lightblue}(↑0.53)} & 82.30 {\scriptsize \color{lightblue}(↓5.06)} & 96.36 {\scriptsize \color{lightblue}(↑5.30)} & 67.13 {\scriptsize \color{lightblue}(↓15.73)} & 79.14 {\scriptsize \color{lightblue}(↓7.63)} \\
& VAF     & 84.06 {\scriptsize \color{lightblue}(↑5.96)} & 83.48 {\scriptsize \color{lightblue}(↑11.77)} & 84.93 {\scriptsize \color{lightblue}(↓7.87)} & 84.20 {\scriptsize \color{lightblue}(↑3.30)} & 87.03 {\scriptsize \color{lightblue}(↓0.33)} & 94.70 {\scriptsize \color{lightblue}(↑3.64)} & 79.73 {\scriptsize \color{lightblue}(↓3.13)} & 86.56 {\scriptsize \color{lightblue}(↓0.21)} \\
& ICT     & 84.62 {\scriptsize \color{lightblue}(↑6.52)} & 85.13 {\scriptsize \color{lightblue}(↑13.42)} & 84.40 {\scriptsize \color{lightblue}(↓8.40)} & 84.76 {\scriptsize \color{lightblue}(↑3.86)} & 88.13 {\scriptsize \color{lightblue}(↑0.77)} & 92.81 {\scriptsize \color{lightblue}(↑1.75)} & 85.00 {\scriptsize \color{lightblue}(↑2.14)} & 88.74 {\scriptsize \color{lightblue}(↑1.97)} \\
& VTI     & 83.33 {\scriptsize \color{lightblue}(↑5.23)} & 82.47 {\scriptsize \color{lightblue}(↑10.76)} & 84.67 {\scriptsize \color{lightblue}(↓8.13)} & 83.55 {\scriptsize \color{lightblue}(↑2.65)} & 87.23 {\scriptsize \color{lightblue}(↓0.13)} & 91.46 {\scriptsize \color{lightblue}(↑0.40)} & 82.13 {\scriptsize \color{lightblue}(↓0.73)} & 86.55 {\scriptsize \color{lightblue}(↓0.22)} \\
& Nullu   & 81.16 {\scriptsize \color{lightblue}(↑3.06)} & 77.00 {\scriptsize \color{lightblue}(↑5.29)} & 88.86 {\scriptsize \color{lightblue}(↓3.94)} & 82.51 {\scriptsize \color{lightblue}(↑1.61)} & 87.03 {\scriptsize \color{lightblue}(↓0.33)} & 90.45 {\scriptsize \color{lightblue}(↓0.61)} & 82.80 {\scriptsize \color{lightblue}(↓0.06)} & 86.46 {\scriptsize \color{lightblue}(↓0.31)} \\
& MESA    & \textbf{84.80} {\scriptsize \color{lightblue}(↑6.70)} & 84.16 {\scriptsize \color{lightblue}(↑12.45)} & 85.73 {\scriptsize \color{lightblue}(↓7.07)} & \textbf{84.94} {\scriptsize \color{lightblue}(↑4.04)} & \textbf{89.77} {\scriptsize \color{lightblue}(↑2.41)} & 93.07 {\scriptsize \color{lightblue}(↑2.01)} & 85.93 {\scriptsize \color{lightblue}(↑3.07)} & \textbf{89.36} {\scriptsize \color{lightblue}(↑2.59)} \\

\midrule

\multirow[c]{8}{*}{Adversarial}
& Vanilla & 67.50 & 61.62 & 92.80 & 74.06 & 79.56 & 77.66 & 83.00 & 80.24 \\
& VCD     & 74.97 {\scriptsize \color{lightblue}(↑7.47)} & 70.01 {\scriptsize \color{lightblue}(↑8.39)} & 87.36 {\scriptsize \color{lightblue}(↓5.44)} & 77.73 {\scriptsize \color{lightblue}(↑3.67)} & 81.27 {\scriptsize \color{lightblue}(↑1.71)} & 77.79 {\scriptsize \color{lightblue}(↑0.13)} & 87.53 {\scriptsize \color{lightblue}(↑4.53)} & 82.38 {\scriptsize \color{lightblue}(↑2.14)} \\
& ICD     & 75.70 {\scriptsize \color{lightblue}(↑8.20)} & 73.43 {\scriptsize \color{lightblue}(↑11.81)} & 80.53 {\scriptsize \color{lightblue}(↓12.27)} & 76.82 {\scriptsize \color{lightblue}(↑2.76)} & 79.17 {\scriptsize \color{lightblue}(↓0.39)} & 86.61 {\scriptsize \color{lightblue}(↑8.95)} & 69.00 {\scriptsize \color{lightblue}(↓14.00)} & 76.81 {\scriptsize \color{lightblue}(↓3.43)} \\
& VAF     & 76.93 {\scriptsize \color{lightblue}(↑9.43)} & 73.21 {\scriptsize \color{lightblue}(↑11.59)} & 84.93 {\scriptsize \color{lightblue}(↓7.87)} & 78.64 {\scriptsize \color{lightblue}(↑4.58)} & 81.06 {\scriptsize \color{lightblue}(↑1.50)} & 85.78 {\scriptsize \color{lightblue}(↑8.12)} & 78.73 {\scriptsize \color{lightblue}(↓4.27)} & 82.10 {\scriptsize \color{lightblue}(↑1.86)} \\
& ICT     & \textbf{78.53} {\scriptsize \color{lightblue}(↑11.03)} & 75.47 {\scriptsize \color{lightblue}(↑13.85)} & 84.53 {\scriptsize \color{lightblue}(↓8.27)} & \textbf{79.74} {\scriptsize \color{lightblue}(↑5.68)} & 81.56 {\scriptsize \color{lightblue}(↑2.00)} & 79.19 {\scriptsize \color{lightblue}(↑1.53)} & 84.11 {\scriptsize \color{lightblue}(↑1.11)} & 81.59 {\scriptsize \color{lightblue}(↑1.35)} \\
& VTI     & 76.77 {\scriptsize \color{lightblue}(↑9.27)} & 73.11 {\scriptsize \color{lightblue}(↑11.49)} & 84.67 {\scriptsize \color{lightblue}(↓8.13)} & 78.47 {\scriptsize \color{lightblue}(↑4.41)} & 80.57 {\scriptsize \color{lightblue}(↑1.01)} & 79.64 {\scriptsize \color{lightblue}(↑1.98)} & 82.13 {\scriptsize \color{lightblue}(↓0.87)} & 80.87 {\scriptsize \color{lightblue}(↑0.63)} \\
& Nullu   & 72.13 {\scriptsize \color{lightblue}(↑4.63)} & 66.58 {\scriptsize \color{lightblue}(↑4.96)} & 88.86 {\scriptsize \color{lightblue}(↓3.94)} & 76.12 {\scriptsize \color{lightblue}(↑2.06)} & 79.30 {\scriptsize \color{lightblue}(↓0.26)} & 77.31 {\scriptsize \color{lightblue}(↓0.35)} & 82.93 {\scriptsize \color{lightblue}(↓0.07)} & 80.02 {\scriptsize \color{lightblue}(↓0.22)} \\
& MESA    & 78.03 {\scriptsize \color{lightblue}(↑10.53)} & 74.29 {\scriptsize \color{lightblue}(↑12.67)} & 85.73 {\scriptsize \color{lightblue}(↓7.07)} & 79.60 {\scriptsize \color{lightblue}(↑5.54)} & \textbf{82.00} {\scriptsize \color{lightblue}(↑2.44)} & 79.67 {\scriptsize \color{lightblue}(↑2.01)} & 85.93 {\scriptsize \color{lightblue}(↑2.93)} & \textbf{82.68} {\scriptsize \color{lightblue}(↑2.44)} \\

\bottomrule
\end{tabular}
\label{tab:pope_aokvqa}
\end{table*}


\begin{table*}[t!]
\centering
\caption{Performance Comparison under Different Settings on GQA with POPE Evaluation}
\setlength{\tabcolsep}{4pt}
\renewcommand{\arraystretch}{0.9}

\begin{tabular}{l|l|cccc|cccc}
\toprule
\multirow{2}{*}{Setting} & \multirow{2}{*}{Method} 
& \multicolumn{4}{c|}{\textbf{LLaVA-v1.5}} 
& \multicolumn{4}{c}{\textbf{Qwen-VL}} \\
\cmidrule(lr){3-6} \cmidrule(lr){7-10}
& 
& Accuracy$\uparrow$ & Precision$\uparrow$ & Recall$\uparrow$ & F1 Score$\uparrow$
& Accuracy$\uparrow$ & Precision$\uparrow$ & Recall$\uparrow$ & F1 Score$\uparrow$ \\
\midrule

\multirow[c]{8}{*}{Random}
& Vanilla & 84.96 & 79.44 & 94.33 & 86.25 & 74.76 & 98.43 & 50.33 & 66.60 \\
& VCD     & 86.65 {\scriptsize \color{lightblue}(↑1.69)} & 84.85 {\scriptsize \color{lightblue}(↑5.41)} & 89.24 {\scriptsize \color{lightblue}(↓5.09)} & 86.99 {\scriptsize \color{lightblue}(↑0.74)} & 85.59 {\scriptsize \color{lightblue}(↑10.83)} & 86.88 {\scriptsize \color{lightblue}(↓11.55)} & 83.84 {\scriptsize \color{lightblue}(↑33.51)} & 85.33 {\scriptsize \color{lightblue}(↑18.73)} \\
& ICD     & 85.90 {\scriptsize \color{lightblue}(↑0.94)} & 89.10 {\scriptsize \color{lightblue}(↑9.66)} & 81.80 {\scriptsize \color{lightblue}(↓12.53)} & 85.29 {\scriptsize \color{lightblue}(↓0.96)} & 86.20 {\scriptsize \color{lightblue}(↑11.44)} & 95.55 {\scriptsize \color{lightblue}(↓2.88)} & 75.93 {\scriptsize \color{lightblue}(↑25.60)} & 84.62 {\scriptsize \color{lightblue}(↑18.02)} \\
& VAF     & 89.00 {\scriptsize \color{lightblue}(↑4.04)} & 91.66 {\scriptsize \color{lightblue}(↑12.22)} & 85.80 {\scriptsize \color{lightblue}(↓8.53)} & 88.63 {\scriptsize \color{lightblue}(↑2.38)} & 86.11 {\scriptsize \color{lightblue}(↑11.35)} & 96.87 {\scriptsize \color{lightblue}(↓1.56)} & 77.53 {\scriptsize \color{lightblue}(↑27.20)} & 86.11 {\scriptsize \color{lightblue}(↑19.51)} \\
& ICT     & 89.26 {\scriptsize \color{lightblue}(↑4.30)} & 92.93 {\scriptsize \color{lightblue}(↑13.49)} & 85.00 {\scriptsize \color{lightblue}(↓9.33)} & 88.78 {\scriptsize \color{lightblue}(↑2.53)} & 85.33 {\scriptsize \color{lightblue}(↑10.57)} & 95.95 {\scriptsize \color{lightblue}(↓2.48)} & 78.13 {\scriptsize \color{lightblue}(↑27.80)} & 86.12 {\scriptsize \color{lightblue}(↑19.52)} \\
& VTI     & 88.37 {\scriptsize \color{lightblue}(↑3.41)} & 90.73 {\scriptsize \color{lightblue}(↑11.29)} & 85.47 {\scriptsize \color{lightblue}(↓8.86)} & 88.02 {\scriptsize \color{lightblue}(↑1.77)} & 84.33 {\scriptsize \color{lightblue}(↑9.57)} & 96.73 {\scriptsize \color{lightblue}(↓1.70)} & 71.07 {\scriptsize \color{lightblue}(↑20.74)} & 81.94 {\scriptsize \color{lightblue}(↑15.34)} \\
& Nullu   & 87.70 {\scriptsize \color{lightblue}(↑2.74)} & 85.90 {\scriptsize \color{lightblue}(↑6.46)} & 90.20 {\scriptsize \color{lightblue}(↓4.13)} & 88.00 {\scriptsize \color{lightblue}(↑1.75)} & 74.73 {\scriptsize \color{lightblue}(↓0.03)} & 98.43 {\scriptsize \color{lightblue}(↑0.00)} & 50.26 {\scriptsize \color{lightblue}(↓0.07)} & 65.55 {\scriptsize \color{lightblue}(↓1.05)} \\
& MESA    & \textbf{89.50} {\scriptsize \color{lightblue}(↑4.54)} & 90.00 {\scriptsize \color{lightblue}(↑10.56)} & 89.00 {\scriptsize \color{lightblue}(↓5.33)} & \textbf{89.45} {\scriptsize \color{lightblue}(↑3.20)} & \textbf{87.97} {\scriptsize \color{lightblue}(↑13.21)} & 97.11 {\scriptsize \color{lightblue}(↓1.32)} & 78.27 {\scriptsize \color{lightblue}(↑27.94)} & \textbf{86.67} {\scriptsize \color{lightblue}(↑20.07)} \\

\midrule

\multirow[c]{8}{*}{Popular}
& Vanilla & 73.30 & 66.40 & 94.33 & 77.93 & 74.76 & 98.18 & 50.46 & 66.66 \\
& VCD     & 80.73 {\scriptsize \color{lightblue}(↑7.43)} & 76.26 {\scriptsize \color{lightblue}(↑9.86)} & 89.24 {\scriptsize \color{lightblue}(↓5.09)} & 82.24 {\scriptsize \color{lightblue}(↑4.31)} & 81.83 {\scriptsize \color{lightblue}(↑7.07)} & 80.45 {\scriptsize \color{lightblue}(↓17.73)} & 84.09 {\scriptsize \color{lightblue}(↑33.63)} & 82.23 {\scriptsize \color{lightblue}(↑15.57)} \\
& ICD     & 78.56 {\scriptsize \color{lightblue}(↑5.26)} & 77.95 {\scriptsize \color{lightblue}(↑11.55)} & 79.66 {\scriptsize \color{lightblue}(↓14.67)} & 78.79 {\scriptsize \color{lightblue}(↑0.86)} & 80.80 {\scriptsize \color{lightblue}(↑6.04)} & 91.77 {\scriptsize \color{lightblue}(↓6.41)} & 67.66 {\scriptsize \color{lightblue}(↑17.20)} & 77.89 {\scriptsize \color{lightblue}(↑11.23)} \\
& VAF     & 81.16 {\scriptsize \color{lightblue}(↑7.86)} & 78.52 {\scriptsize \color{lightblue}(↑12.12)} & 85.80 {\scriptsize \color{lightblue}(↓8.53)} & 82.00 {\scriptsize \color{lightblue}(↑4.07)} & 85.43 {\scriptsize \color{lightblue}(↑10.67)} & 92.61 {\scriptsize \color{lightblue}(↓5.57)} & 72.53 {\scriptsize \color{lightblue}(↑22.07)} & 81.35 {\scriptsize \color{lightblue}(↑14.69)} \\
& ICT     & 82.86 {\scriptsize \color{lightblue}(↑9.56)} & 82.51 {\scriptsize \color{lightblue}(↑16.11)} & 84.93 {\scriptsize \color{lightblue}(↓9.40)} & 83.70 {\scriptsize \color{lightblue}(↑5.77)} & 85.31 {\scriptsize \color{lightblue}(↑10.55)} & 91.59 {\scriptsize \color{lightblue}(↓6.59)} & 77.86 {\scriptsize \color{lightblue}(↑27.40)} & 84.15 {\scriptsize \color{lightblue}(↑17.49)} \\
& VTI     & 81.73 {\scriptsize \color{lightblue}(↑8.43)} & 79.53 {\scriptsize \color{lightblue}(↑13.13)} & 85.47 {\scriptsize \color{lightblue}(↓8.86)} & 82.39 {\scriptsize \color{lightblue}(↑4.46)} & 83.67 {\scriptsize \color{lightblue}(↑8.91)} & 95.01 {\scriptsize \color{lightblue}(↓3.17)} & 71.07 {\scriptsize \color{lightblue}(↑20.61)} & 81.31 {\scriptsize \color{lightblue}(↑14.65)} \\
& Nullu   & 77.63 {\scriptsize \color{lightblue}(↑4.33)} & 72.08 {\scriptsize \color{lightblue}(↑5.68)} & 90.20 {\scriptsize \color{lightblue}(↓4.13)} & 80.13 {\scriptsize \color{lightblue}(↑2.20)} & 74.71 {\scriptsize \color{lightblue}(↓0.05)} & 98.17 {\scriptsize \color{lightblue}(↓0.01)} & 50.33 {\scriptsize \color{lightblue}(↓0.13)} & 66.54 {\scriptsize \color{lightblue}(↓0.12)} \\
& MESA    & \textbf{82.93} {\scriptsize \color{lightblue}(↑9.63)} & 79.37 {\scriptsize \color{lightblue}(↑12.97)} & 89.00 {\scriptsize \color{lightblue}(↓5.33)} & \textbf{83.91} {\scriptsize \color{lightblue}(↑5.98)} & \textbf{86.07} {\scriptsize \color{lightblue}(↑11.31)} & 92.73 {\scriptsize \color{lightblue}(↓5.45)} & 78.27 {\scriptsize \color{lightblue}(↑27.81)} & \textbf{84.89} {\scriptsize \color{lightblue}(↑18.23)} \\

\midrule

\multirow[c]{8}{*}{Adversarial}
& Vanilla & 67.40 & 61.30 & 94.33 & 74.31 & 73.53 & 93.90 & 50.33 & 65.53 \\
& VCD     & 76.09 {\scriptsize \color{lightblue}(↑8.69)} & 70.83 {\scriptsize \color{lightblue}(↑9.53)} & 88.75 {\scriptsize \color{lightblue}(↓5.58)} & 78.78 {\scriptsize \color{lightblue}(↑4.47)} & 80.01 {\scriptsize \color{lightblue}(↑6.48)} & 77.86 {\scriptsize \color{lightblue}(↓16.04)} & 83.85 {\scriptsize \color{lightblue}(↑33.52)} & 80.75 {\scriptsize \color{lightblue}(↑15.22)} \\
& ICD     & 75.90 {\scriptsize \color{lightblue}(↑8.50)} & 73.84 {\scriptsize \color{lightblue}(↑12.54)} & 80.20 {\scriptsize \color{lightblue}(↓14.13)} & 76.89 {\scriptsize \color{lightblue}(↑2.58)} & 81.00 {\scriptsize \color{lightblue}(↑7.47)} & 84.24 {\scriptsize \color{lightblue}(↓9.66)} & 76.27 {\scriptsize \color{lightblue}(↑25.94)} & 80.06 {\scriptsize \color{lightblue}(↑14.53)} \\
& VAF     & 79.10 {\scriptsize \color{lightblue}(↑11.70)} & 75.66 {\scriptsize \color{lightblue}(↑14.36)} & 85.80 {\scriptsize \color{lightblue}(↓8.53)} & 80.41 {\scriptsize \color{lightblue}(↑6.10)} & 80.03 {\scriptsize \color{lightblue}(↑6.50)} & 83.30 {\scriptsize \color{lightblue}(↓10.60)} & 76.53 {\scriptsize \color{lightblue}(↑26.20)} & 79.75 {\scriptsize \color{lightblue}(↑14.22)} \\
& ICT     & 79.13 {\scriptsize \color{lightblue}(↑11.73)} & 78.17 {\scriptsize \color{lightblue}(↑16.87)} & 84.16 {\scriptsize \color{lightblue}(↓10.17)} & 81.07 {\scriptsize \color{lightblue}(↑6.76)} & 81.61 {\scriptsize \color{lightblue}(↑8.08)} & 84.61 {\scriptsize \color{lightblue}(↓9.29)} & 77.21 {\scriptsize \color{lightblue}(↑26.88)} & 80.75 {\scriptsize \color{lightblue}(↑15.22)} \\
& VTI     & 78.33 {\scriptsize \color{lightblue}(↑10.93)} & 74.80 {\scriptsize \color{lightblue}(↑13.50)} & 85.47 {\scriptsize \color{lightblue}(↓8.86)} & 79.78 {\scriptsize \color{lightblue}(↑5.47)} & 80.47 {\scriptsize \color{lightblue}(↑6.94)} & 87.52 {\scriptsize \color{lightblue}(↓6.38)} & 71.07 {\scriptsize \color{lightblue}(↑20.74)} & 78.44 {\scriptsize \color{lightblue}(↑12.91)} \\
& Nullu   & 72.10 {\scriptsize \color{lightblue}(↑4.70)} & 66.22 {\scriptsize \color{lightblue}(↑4.92)} & 90.20 {\scriptsize \color{lightblue}(↓4.13)} & 76.37 {\scriptsize \color{lightblue}(↑2.06)} & 73.61 {\scriptsize \color{lightblue}(↑0.08)} & 94.13 {\scriptsize \color{lightblue}(↑0.23)} & 50.33 {\scriptsize \color{lightblue}(↑0.00)} & 65.59 {\scriptsize \color{lightblue}(↑0.06)} \\
\rowcolor{gray!15}
& MESA    & \textbf{79.60} {\scriptsize \color{lightblue}(↑12.20)} & 74.92 {\scriptsize \color{lightblue}(↑13.62)} & 89.00 {\scriptsize \color{lightblue}(↓5.33)} & \textbf{81.35} {\scriptsize \color{lightblue}(↑7.04)} & \textbf{82.73} {\scriptsize \color{lightblue}(↑9.20)} & 85.94 {\scriptsize \color{lightblue}(↓7.96)} & 78.27 {\scriptsize \color{lightblue}(↑27.94)} & \textbf{81.93} {\scriptsize \color{lightblue}(↑16.40)} \\

\bottomrule
\end{tabular}
\label{tab:pope_gqa}
\end{table*}

\begin{table*}[t!]
\centering
\caption{Performance comparison on generative and discriminative tasks with AMBER Evaluation.}
\setlength{\tabcolsep}{6pt}
\renewcommand{\arraystretch}{0.9}
\begin{tabular}{c|cccc|cccc}
\toprule
\multirow{2}{*}{Method} 
& \multicolumn{4}{c|}{Generative Task} 
& \multicolumn{4}{c}{Discriminative Task} \\
\cmidrule(lr){2-5} \cmidrule(lr){6-9}
& CHAIR$\downarrow$ & Cover$\uparrow$ & Hallucination$\downarrow$ & Cognition$\downarrow$
& Accuracy$\uparrow$ & Precision$\uparrow$ & Recall$\uparrow$ & F1 Score$\uparrow$ \\
\midrule

Vanilla 
& 10.6 & 50.9 & 36.4 & 4.1 
& 71.4 & 81.9 & 73.1 & 77.2 \\

VCD 
& 8.9 {\scriptsize \color{lightblue}(↓1.70)} & \textbf{51.8} {\scriptsize \color{lightblue}(↑0.90)} & 41.9 {\scriptsize \color{lightblue}(↑5.50)} & 4.3 {\scriptsize \color{lightblue}(↑0.20)} 
& 73.9 {\scriptsize \color{lightblue}(↑2.50)} & 83.8 {\scriptsize \color{lightblue}(↑1.90)} & 75.2 {\scriptsize \color{lightblue}(↑2.10)} & 79.3 {\scriptsize \color{lightblue}(↑2.10)} \\

ICD 
& 9.3 {\scriptsize \color{lightblue}(↓1.30)} 
& 50.5 {\scriptsize \color{lightblue}(↓0.40)} 
& 41.3 {\scriptsize \color{lightblue}(↑4.90)} 
& 4.9 {\scriptsize \color{lightblue}(↑0.80)} 
& 74.1 {\scriptsize \color{lightblue}(↑2.70)} & 73.2 {\scriptsize \color{lightblue}(↓8.70)} & 96.2 {\scriptsize \color{lightblue}(↑23.10)} & 83.1 {\scriptsize \color{lightblue}(↑5.90)} \\

VAF 
& 10.9 {\scriptsize \color{lightblue}(↑0.30)} & 33.0 {\scriptsize \color{lightblue}(↓17.90)} & 28.1 {\scriptsize \color{lightblue}(↓8.30)} & 3.7 {\scriptsize \color{lightblue}(↓0.40)} 
& 69.8 {\scriptsize \color{lightblue}(↓1.60)} & 94.8 {\scriptsize \color{lightblue}(↑12.90)} & 57.6 {\scriptsize \color{lightblue}(↓15.50)} & 71.7 {\scriptsize \color{lightblue}(↓5.50)} \\

ICT 
& 7.0 {\scriptsize \color{lightblue}(↓3.60)} & 48.0 {\scriptsize \color{lightblue}(↓2.90)} & 28.4 {\scriptsize \color{lightblue}(↓8.00)} & 3.5 {\scriptsize \color{lightblue}(↓0.60)} 
& 79.5 {\scriptsize \color{lightblue}(↑8.10)} & 84.7 {\scriptsize \color{lightblue}(↑2.80)} & 84.4 {\scriptsize \color{lightblue}(↑11.30)} & 84.5 {\scriptsize \color{lightblue}(↑7.30)} \\

VTI 
& 6.7 {\scriptsize \color{lightblue}(↓3.90)} & 47.6 {\scriptsize \color{lightblue}(↓3.30)} & 28.1 {\scriptsize \color{lightblue}(↓8.30)} & 3.6 {\scriptsize \color{lightblue}(↓0.50)} 
& 79.1 {\scriptsize \color{lightblue}(↑7.70)} & 82.1 {\scriptsize \color{lightblue}(↑0.20)} & 87.6 {\scriptsize \color{lightblue}(↑14.50)} & 84.8 {\scriptsize \color{lightblue}(↑7.60)} \\

Nullu 
& 7.4 {\scriptsize \color{lightblue}(↓3.20)} & 48.2 {\scriptsize \color{lightblue}(↓2.70)} & 30.5 {\scriptsize \color{lightblue}(↓5.90)} & 3.7 {\scriptsize \color{lightblue}(↓0.40)} 
& 81.5 {\scriptsize \color{lightblue}(↑10.10)} & 87.4 {\scriptsize \color{lightblue}(↑5.50)} & 84.2 {\scriptsize \color{lightblue}(↑11.10)} & 85.8 {\scriptsize \color{lightblue}(↑8.60)} \\

\rowcolor{gray!15}
MESA 
& \textbf{6.4} {\scriptsize \color{lightblue}(↓4.20)} & 50.4 {\scriptsize \color{lightblue}(↓0.50)} & \textbf{27.4} {\scriptsize \color{lightblue}(↓9.00)} & \textbf{3.2} {\scriptsize \color{lightblue}(↓0.90)} 
& \textbf{84.3} {\scriptsize \color{lightblue}(↑12.90)} & 90.3 {\scriptsize \color{lightblue}(↑8.40)} & 85.4 {\scriptsize \color{lightblue}(↑12.30)} & \textbf{87.8} {\scriptsize \color{lightblue}(↑10.60)} \\

\bottomrule
\end{tabular}
\label{tab:amber_full}
\end{table*}

\section{More Experimental Results}
\subsection{Results on POPE}
As demonstrated by the comprehensive experimental results in ~\autoref{tab:pope_mscoco}, ~\autoref{tab:pope_aokvqa}, and ~\autoref{tab:pope_gqa}, most existing methods mitigate hallucinations primarily by improving precision, but at the cost of a noticeable drop in recall, indicating a clear precision--recall trade-off.
For instance, methods such as ICD and VAF substantially increase precision, while significantly suppressing recall, which in turn limits their overall F1 Score performance.
In contrast, MESA achieves a more balanced behavior across all settings.
While consistently improving precision, it maintains relatively stable recall compared to other baselines, leading to superior F1 scores.
This advantage is particularly evident under the Adversarial setting, where MESA delivers the best overall performance, demonstrating its robustness in challenging scenarios.
Overall, these results suggest that MESA effectively suppresses hallucinations without over-pruning valid predictions, thereby achieving a better trade-off between precision and recall.

\subsection{Results on AMBER}
\autoref{tab:amber_full} presents the complete results on the AMBER benchmark using LLaVA-v1.5.
On the generative side, methods such as ICT, VTI, and Nullu reduce CHAIR and hallucination scores, but often at the cost of coverage.
VAF further demonstrates this issue, achieving strong hallucination reduction but significantly harming coverage, while VCD improves coverage yet increases hallucination.
On the discriminative side, most methods improve accuracy or recall, but introduce a precision--recall imbalance.
For example, ICD yields high recall with low precision, whereas VAF shows the opposite trend.
In contrast, our method (Ours) achieves the best overall balance.
It obtains the lowest CHAIR and hallucination scores while maintaining competitive coverage and the best cognition score.
Meanwhile, it achieves the highest accuracy and F1 score, demonstrating consistent gains without sacrificing precision or recall.
These results indicate that our method effectively suppresses hallucinations while preserving overall generation quality and task performance.

\begin{figure}[!t]
\centering
\includegraphics[width=0.7\linewidth]{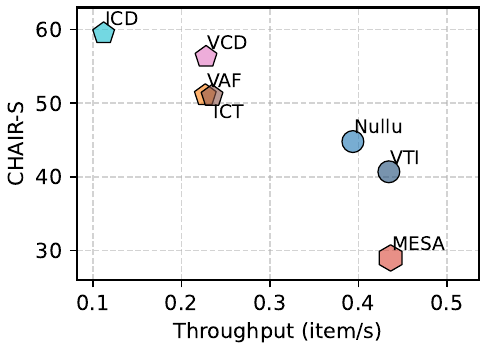}
\caption{The throughput (tested on NVIDIA A800) v.s. $\mathrm{CHAIR}_S$ on CHAIR benchmark.}
\label{fig:cost}
\end{figure}

\subsection{Case Study}
\begin{figure*}[!t]
\centering
\makebox[\textwidth][c]{%
    \includegraphics[width=1.2\linewidth]{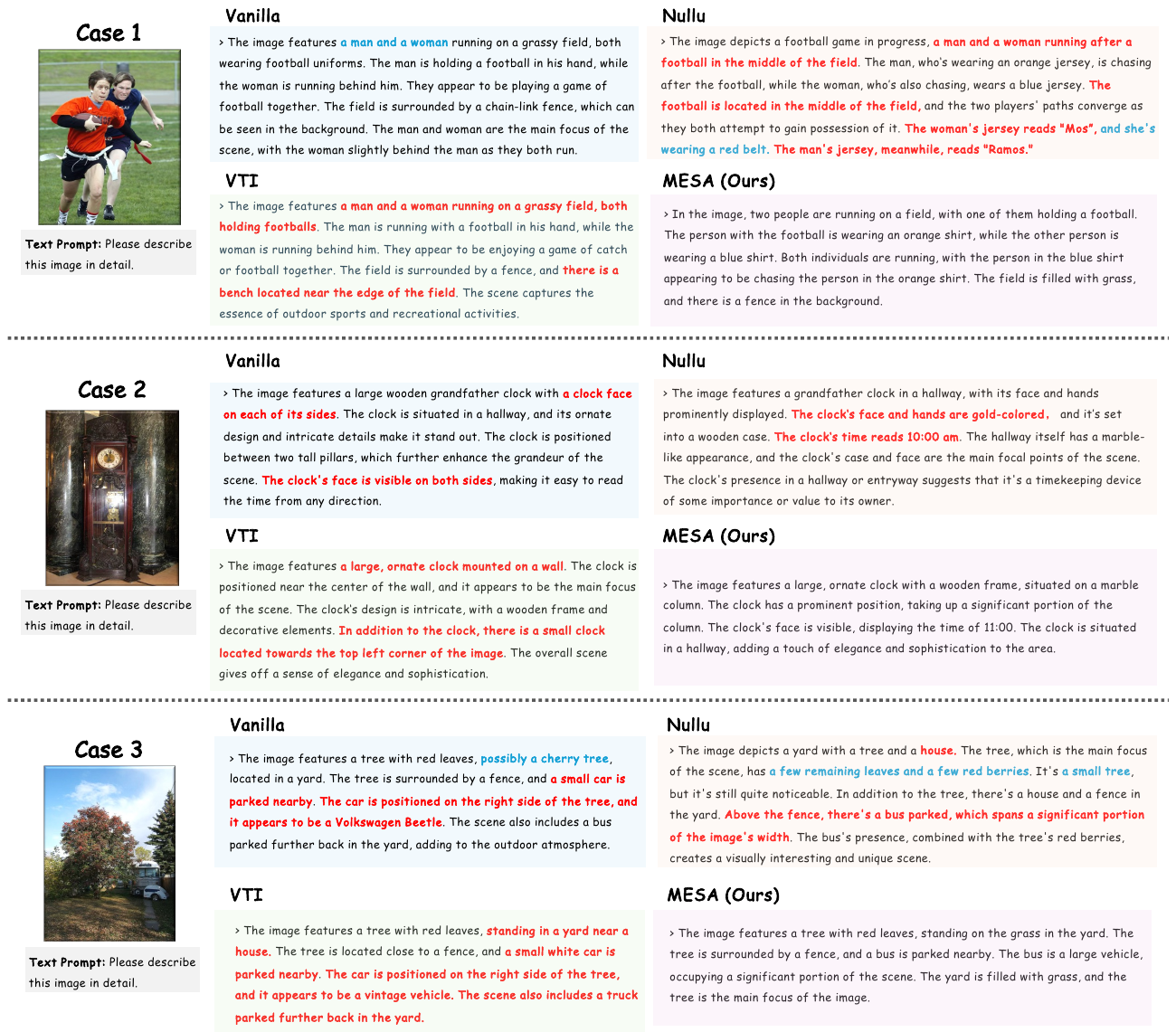}
}
\vspace{-10mm} 
\caption{Illustration of hallucination correction comparisons. 
Text highlighted in red bold denotes explicit hallucinations or factual errors, whereas blue bold highlights indicate milder but potentially questionable over-inferences.}
\label{fig:case_all}
\end{figure*}

\autoref{fig:case_all} present three case studies on LLaVA-1.5-7B using identical prompts (\textit{"Please describe this image in detail."}) and images from MSCOCO. 
Text highlighted in red bold indicates explicit hallucinations or factual errors, while blue bold denotes weaker but potentially questionable over-inferences.

\circled{1} \ \textbf{Case 1.}
In this example, our proposed MESA most accurately captures the core action relationship in the image, correctly identifying that “a foreground player is advancing with the ball while another follows behind,” without introducing additional objects or details. 
In contrast, Nullu fabricates “another ball at the center of the field” and incorrectly recognizes text on the jersey. VTI misinterprets the scene as “both individuals holding a ball” and further introduces a non-existent bench. 
Moreover, all methods except ours assume the subjects are a man and a woman, whereas the gender cannot be reliably determined from the image alone.

\circled{2} \ \textbf{Case 2.}
Although this image depicts a relatively simple static indoor scene, competing methods still exhibit clear hallucinations. 
Vanilla incorrectly describes “clock faces on both sides,” Nullu assigns an inaccurate specific time, and VTI misclassifies the object as “a wall-mounted clock” while additionally introducing a non-existent smaller clock. 
By comparison, our method correctly captures the key entities and their spatial arrangement, achieving substantially higher fidelity.

\circled{3} \ \textbf{Case 3.}
This case highlights differences under a scenario where the main subject is clear but background object categories are prone to misclassification. 
Both Vanilla and VTI misidentify the large vehicle and trailer on the right as a car, while VTI further describes a background vehicle as a truck. 
Nullu exhibits significant deviations in the scale and detail of the tree and bus, and also incorrectly states that the bus almost spans the entire width of the image. 
In contrast, our method preserves the two most salient visual facts, namely the tree in the yard and the large bus nearby, resulting in a more faithful overall description.

\subsection{Analysis of rank in PCA}
In this paper, we compute the layer-wise difference $\Delta h^{(l)} = h^{(l)} - \tilde{h}^{(l)}$ to characterize perturbation-induced shifts in the representation space. These difference vectors capture how the model’s internal activations respond to controlled perturbations and thus serve as a basis for identifying corrective directions. We then apply Principal Component Analysis (PCA) to $\Delta h^{(l)}$ aggregated across samples, extracting the dominant principal components as candidate steering directions.~\autoref{tab:abla_rank} presents the effect of varying the PCA rank on LLaVA-v1.5 7B evaluated on the CHAIR benchmark. The results show that rank = 4 achieves slightly better performance than other configurations, indicating that incorporating multiple principal directions can provide marginal gains. However, the overall performance differences across ranks are relatively small, suggesting that the dominant variance is already well captured by the leading component. Therefore, we adopt rank = 1 in our method, as it offers a favorable trade-off between effectiveness and computational efficiency while maintaining stable performance.

\subsection{Analysis of top-$m$}
\begin{figure*}[t]
\centering
\begin{subfigure}{0.23\textwidth}
    \centering
    \includegraphics[width=\linewidth]{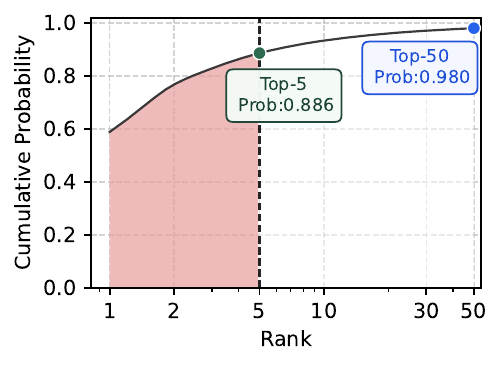}
    \caption{Cumulative probability mass over token ranks for tokens generated by the original model.}
    \label{fig:orig_top50}
\end{subfigure}
\hfill
\begin{subfigure}{0.23\textwidth}
    \centering
    \includegraphics[width=\linewidth]{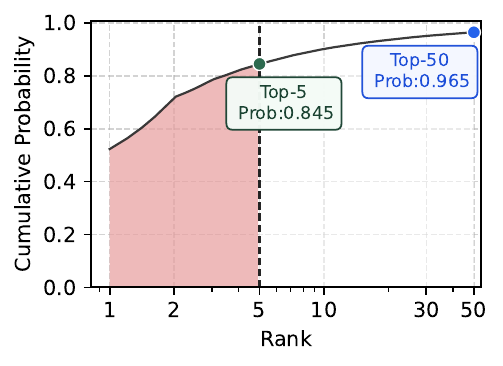}
    \caption{Cumulative probability mass over token ranks under Jigsaw-induced hallucination.}
    \label{fig:jigsaw_top50}
\end{subfigure}
\hfill
\begin{subfigure}{0.23\textwidth}
    \centering
    \includegraphics[width=\linewidth]{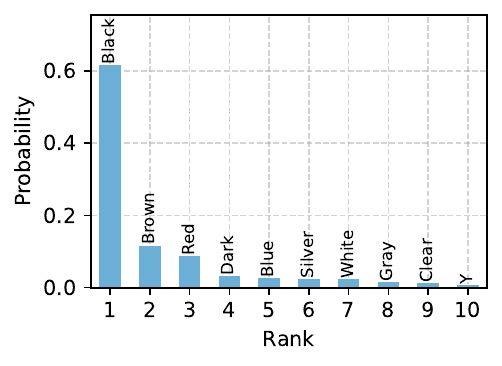}
    \caption{Example of top-ranked tokens and their probabilities from the original model.}
    \label{fig:orig_topk_tokens}
\end{subfigure}
\hfill
\begin{subfigure}{0.23\textwidth}
    \centering
    \includegraphics[width=\linewidth]{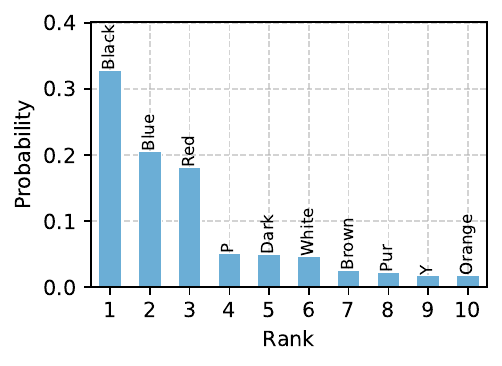}
    \caption{Example of top-ranked tokens and probabilities under Jigsaw-induced hallucination.}
    \label{fig:jigsaw_topk_tokens}
\end{subfigure}
\caption{
Comparison of token probability distributions and representative examples under normal and hallucination-inducing conditions.
(a)-(b): The cumulative probability mass over token ranks shows a consistent concentration pattern.
(c)-(d): Token-level examples illustrate that Jigsaw-induced hallucination alters the probability distribution of top-ranked tokens.
}
\label{fig:generation_behavior_mesa}
\end{figure*}

In our method, the hallucination objective $\mathcal{L}_{\text{hall}}$ is computed over the top-50 tokens, while the preservation objective $\mathcal{L}_{\text{preserve}}$ excludes the highest-probability tokens (e.g., top-$m$ with $m=5$) to avoid over-constraining dominant tokens. In this section, we analyze the rationale behind the choice of top-$m$ and top-50.
First, we examine the cumulative probability mass over token ranks for both the original model and under Jigsaw-induced hallucination. As shown in \autoref{fig:orig_top50} and \autoref{fig:jigsaw_top50}, the probability distribution exhibits a highly concentrated structure in both cases: the top-5 tokens account for approximately 84.5\% and 88.6\% of the total probability mass, respectively, while the top-50 tokens cover more than 96\% and 98\%. These results indicate that the top-50 tokens already capture the vast majority of the probability mass, justifying the use of $\mathcal{L}_{\text{hall}}$ over this subset.
In addition, as shown in \autoref{fig:orig_topk_tokens} and \autoref{fig:jigsaw_topk_tokens}, Jigsaw-induced hallucination leads to noticeable changes in both the probabilities and the ranking of top tokens. This suggests that hallucination can be effectively induced by redistributing probability among top-ranked tokens. At the same time, key hallucinated tokens predominantly appear within this high-probability region. Therefore, it is important to preserve the remaining part of the distribution to maintain semantic consistency, which is enforced by $\mathcal{L}_{\text{preserve}}$.

\subsection{Analysis of Cost}
We measure inference efficiency using throughput (item/s), defined as $\frac{N}{T}$, where $N$ is the number of samples and $T$ is the total inference time. Each item denotes a full generation instance with decoding, reflecting end-to-end efficiency.
As shown in~\autoref{fig:cost}, methods exhibit different trade-offs between hallucination and efficiency.
ICD achieves the lowest throughput despite strong hallucination mitigation, indicating high computational overhead.
VCD, VAF, and ICT provide moderate performance in both metrics.
VTI and Nullu achieve the highest throughput, but with relatively worse hallucination scores.
In comparison, MESA attains a low $\mathrm{CHAIR}_S$ (29.0) while maintaining high throughput, demonstrating strong efficiency without sacrificing generation quality.

\begin{table}[t!]
\centering
\caption{Performance under different rank settings.}
\setlength{\tabcolsep}{4pt}
\renewcommand{\arraystretch}{0.9}
\begin{tabular}{c|cccc}
\toprule
Rank & $\mathrm{CHAIR}_S$\scriptsize$\downarrow$ 
& $\mathrm{CHAIR}_I$\scriptsize$\downarrow$ 
& Recall\scriptsize$\uparrow$ 
& Avg. length\scriptsize$\uparrow$ \\
\midrule
1 & 31.0 & 8.6 & 75.8 & 93.5 \\
2 & 30.8 & 8.6 & 75.8 & 93.4 \\
3 & 31.6 & 8.7 & 75.8 & 93.5 \\
4 & 30.1 & 8.2 & 76.2 & 93.4 \\
5 & 31.2 & 8.5 & 76.1 & 93.7 \\
\bottomrule
\end{tabular}
\label{tab:abla_rank}
\end{table}

\end{document}